\documentclass[sigconf]{acmart}

\usepackage{booktabs} 
\usepackage{times}
\usepackage{helvet}
\usepackage{courier}
\usepackage{color}
\usepackage{amsmath,bm}
\usepackage{array}
\usepackage{multirow}
\usepackage{multicol} 
\usepackage{xfrac}
\usepackage{algorithmic}
\usepackage[inoutnumbered,linesnumbered,ruled,vlined]{algorithm2e}
\usepackage{amsfonts}
\usepackage{graphicx}
\usepackage{mathrsfs}
\usepackage{epsfig}
\usepackage{epstopdf}
\usepackage{subfigure}
\usepackage{enumerate}
\usepackage{helvet}
\usepackage{url}
\usepackage{amsmath,amssymb,amsthm}
\usepackage{bbding}
\usepackage{pifont}
\usepackage{rotating}
\usepackage{mathtools}
\usepackage{enumitem}

\usepackage{balance}
\usepackage{caption}
\usepackage{footnote}
\usepackage{hyperref}
\makesavenoteenv{tabular}
\makesavenoteenv{table}

\setcopyright{rightsretained}

\setcopyright{acmcopyright} 

\begin{document}
\title{Correcting Knowledge Base Assertions}
\author{Jiaoyan Chen}
\affiliation{%
  \institution{University of Oxford}
}
\email{jiaoyan.chen@cs.ox.ac.uk}

\author{Xi Chen}
\affiliation{%
  \institution{Jarvis Lab Tencent}
}
\email{jasonxchen@tencent.com}

\author{Ian Horrocks}
\affiliation{%
  \institution{University of Oxford}
}
\email{ian.horrocks@cs.ox.ac.uk}

\author{Ernesto Jimenez-Ruiz}
\affiliation{%
  \institution{City, University of London\\ University of Oslo}
}
\email{ernesto.jimenez-ruiz@city.ac.uk}
\author{Erik B. Myklebus}
\affiliation{%
  \institution{Norwegian Institute for Water Research\\ University of Oslo}
}
\email{erik.b.myklebust@niva.no}

\begin{abstract}
The usefulness and usability of knowledge bases (KBs) is often limited by quality issues.
One common issue is the presence of erroneous assertions, often caused by lexical or
semantic confusion.
We study the problem of correcting such assertions, 
and present a general correction framework which combines lexical matching, semantic embedding, soft constraint mining and semantic consistency checking.
The framework is evaluated using DBpedia and an enterprise medical KB.
\end{abstract}

%
%

\keywords{Knowledge Base Quality, Assertion Correction, Semantic Embedding, Constraint Mining, Consistency Checking}

\maketitle

\section{Introduction}		
Knowledge bases (KBs) such as Wikidata \cite{vrandevcic2014wikidata} and DBpedia \cite{auer2007dbpedia} are playing an increasingly important role 
in applications such as search engines, question answering, common sense reasoning, data integration and machine learning.
However, they still suffer from various quality issues, including constraint violations and erroneous assertions \cite{farber2018linked,paulheim2017knowledge}, that negatively impact their usefulness and usability.
These may be due to the knowledge itself (e.g., the core knowledge source of DBpedia, Wikipedia, is estimated to have an error rate of $2.8\%$ \cite{weaver2006quantifying}), 
or may be introduced by the knowledge extraction process.

Existing work on KB quality issues covers not only error detection and assessment, 
but also quality improvement via completion, canonicalizaiton and so on \cite{paulheim2017knowledge}.
Regarding error detection, erroneous assertions can be detected by various methods, 
including consistency checking with defined, mined or external constraints \cite{topper2012dbpedia,paulheim2015serving,knublauch2017shapes},
prediction by machine learning or statistics methods \cite{melo2017detection,paulheim2014improving,debattista2016preliminary}, and
evaluation by query templates \cite{kontokostas2014test}; see Section \ref{sec:rw_av} for a more details.
However, the detected erroneous assertions are often eliminated \cite{ngomo2014unsupervised,de2013not},
and few robust methods have been developed to correct them.

Lertvittayakumjorn et al.~\cite{lertvittayakumjorn2017correcting} 
and Melo et al.~\cite{melo2017approach} found that most erroneous assertions are due to confusion or lexical similarity leading to entity misuse; for example confusion between
\textit{Manchester\_United} and \textit{Manchester\_City}, two football clubs based in Manchester, UK, can lead to facts about \textit{Manchester\_United} being incorrectly asserted about \textit{Manchester\_City}.
Such errors are common in not only those general KBs like DBpedia and Wikidata but also those domain KBs like the medical KB used in our evaluation.
Both studies proposed to find an entity to replace either the subject or the object of an erroneous assertion;
however, subject replacement used a simple graph metric and keyword matching, which fails to capture the contextual semantics of the assertion,
while object replacement relies on Wikipedia disambiguation pages, which may be inaccessible or non-existent, and again fail to capture contextual semantics.

Other work has focused on quality improvement, for example by \emph{canonicalizing} assertions whose objects are literals that represent entities (i.e., entity mentions); for example, the literal object in the assertion $\langle$\textit{Yangtze\_River}, \textit{passesArea}, \textit{``three gorges district''}$\rangle$.
Replacing this literal with the entity \textit{Three\_Gorges\_Reservoir\_Region} enriches the semantics of the assertion, which can improve query answering.
Such literal assertions are pervasive in wiki-based KBs such as DBPedia \cite{auer2007dbpedia} and Zhishi.me \cite{niu2011zhishi}, and in open KBs extracted from text;
they may also be introduced when two KBs are aligned or when a KB evolves.
According to the  statistics in \cite{gunaratna2016gleaning}, DBpedia (ca.\ 2016) included over 105,000 such assertions using the property \textit{dbp:location} alone. 
Current methods can predict the type of the entity represented by the literal \cite{gunaratna2016gleaning}, which is useful for creating a new entity,
and can sometimes identify candidate entities in the KB \cite{chen2019canonicalizing},
but they do not propose a general correction method; see Section \ref{sec:rw_kbc} for a more details.


In this paper, we propose a method for correcting assertions whose objects are either erroneous entities or literals.
To this end, we have developed a general framework that exploits related entity estimation, link prediction and constraint-based validation, as shown in Figure~\ref{fig:framework}.
Given a set of target assertions (i.e., assertions that have been identified as erroneous), 
it uses semantic relatedness to identify candidate entities for substitution, extracts a multi-relational graph from the KB (sub-graph) that can model the context of the target assertions, 
and learns a link prediction model using both semantic embeddings and observed features.
The model predicts the assertion likelihood for each candidate substitution,
and filters out those that lead to unlikely assertions.
The framework further verifies the candidate substitutions by checking their consistency w.r.t.\ property range and cardinality constraints mined from the global KB.
The framework finally makes a correction decision, returning a corrected assertion or reporting failure if no likely correction can be identified.

Briefly this paper makes the following main contributions:
\begin{itemize}
\item It proposes a general framework that can correct both erroneous entity assertions and literal assertions;
\item It utilizes both semantic embeddings and observed features to capture the local context used for correction prediction, with a sub-graph extracted for higher efficiency; 
\item It complements the prediction with consistency checking against ``soft'' property constraints mined from the global KB;
\item It evaluates the framework with erroneous entity assertions from a medical KB and literal assertions from DBpedia. 
\end{itemize}


\section{Related Work}\label{sec:rel}
We survey related work in three areas: assertion validation which includes erroneous assertion detection and link prediction with semantic embeddings and observed features, canonicalization, and assertion correction.

\subsection{Assertion Validation}\label{sec:rw_av}
In investigating KB quality, the validity of assertions is clearly an important consideration. 
One way to identify likely invalid assertions is to check consistency against logical constraints or rules.
Explicitly stated KB constraints can be directly used, but these are often weak or even non-existent.
Thus, before using the DBpedia ontology to validate assertions,
Topper et al.~\cite{topper2012dbpedia} enriched it with class disjointness, and property domain and range costraints, all derived via statistical analysis.
Similarly, Paulheim and Gangemi \cite{paulheim2015serving} enriched the ontology via alignment with the DOLCE-Zero foundational ontology.
Various constraint and rule languages, including the Shapes Constraint Language (SHACL) \cite{knublauch2017shapes}, Rule-Based Web Logics \cite{arndt2017using} and SPARQL query templates \cite{kontokostas2014test},
have also been proposed so that external knowledge can be encoded and applied in reasoning for assertion validation.

With the development of machine learning, 
various feature extraction and semantic embedding methods
have been proposed to encode the semantics of entities and relations into vectors \cite{wang2017knowledge}.
%
The observed features are typically indicators (e.g., paths) extracted for a specific prediction problem. 
They often work together with other learning and prediction algorithms, including supervised classification (e.g., PaTyBRED \cite{melo2017approach}), 
autoencoder (e.g., RDF2Vec \cite{ristoski2016rdf2vec}),
statistical distribution estimation (e.g., SDValidate \cite{paulheim2014improving}) and so on.
PaTyBRED and SDValidate directly detect erroneous assertions, 
while RDF2Vec utilizes graph paths to learn intermediate entity representations that can be further used to validate assertions via supervised classification.

In contrast to observed features, which often rely on ad-hoc feature engineering,
semantic embeddings (vectors) can be learned by minimizing an overall loss with a score function for modeling the assertion's likelihood.
They can be directly used to estimate the assertion likelihood with the score function.
State-of-the-art methods and implementations include DistMult \cite{yang2015embedding}, TransE \cite{bordes2013translating}, Neural Tensor Network \cite{socher2013reasoning}, IterE \cite{zhang2019iteratively}, OpenKE \cite{han2018openke} and so on.
They can also be combined with algorithms such as outlier detection \cite{debattista2016preliminary} and supervised classification \cite{myklebust2019knowledge} to deal with assertion validation in specific contexts.

On the one hand, the aforementioned methods were mostly developed for KB completion and erroneous assertion detection, and
few have been applied in assertion correction, especially the semantic embedding methods.
On the other hand, they suffer from various shortcomings that limit their application.
Consistency checking depends on domain knowledge of a specific task for constraint and rule definition, 
while the mined constraints and rules are often weak in modeling local context for disambiguation. 
Semantic embedding methods are good at modeling contextual semantics in a vector space,
but are computationally expensive when learning from large KBs \cite{omran2018scalable} 
and suffer from low robustness when dealing with real world KBs that are often noisy and sparse \cite{pujara2017sparsity}.

\subsection{Canonicalization}\label{sec:rw_kbc}
Recent work on KB canonicalization is relevant to related entity estimation in our setting.
Some of this work focuses on the clustering and disambiguation of entity mentions in an open KB extracted from textual data \cite{galarraga2014canonicalizing,wu2018towards,vashishth2018cesi};
CESI \cite{vashishth2018cesi}, for example, utilizes side information (e.g., WordNet), semantic embedding and clustering to identify equivalent entity mentions.
However, these methods cannot be directly applied in our correction framework 
as they focus on equality while we aim at estimating relatedness, especially for assertions with erroneous objects.
The contexts are also different as, unlike entity mentions, literals have no neighbourhood information (e.g., relationships with other entities) that can be utilized.

Chen et al.~\cite{chen2019canonicalizing} and Gunaratna et al. \cite{gunaratna2016gleaning} aimed at the canonicalization of literal objects used in assertions with DBpedia object properties (whose objects should be entities).
Instead of correcting the literal with an existing entity, they focus on the typing of the entity that the literal represents, which is helpful when a new entity needs to be created for replacement.
Although the work in \cite{chen2019canonicalizing} also tried to identify an existing entity to substitute the literal, their approach suffers from a number of limitations:
the predicted type is used as a constraint for filtering, which is not a
robust and general correction method;
the related entity estimation is ad-hoc and DBpedia specific;
and the type prediction itself only uses entity and property labels, without any other contextual semantics.

\subsection{Assertion Correction}\label{sec:rw_ac}
We focus on recent studies concerning the automatic correction of erroneous assertions.
Some are KB specific.
For example, Dimou et al.~\cite{dimou2015assessing} refined the mappings between Wikipedia data and DBpedia knowledge such that some errors can be corrected during DBpedia construction,
while Pellissier et al.~\cite{pellissier2019learning} mined correction rules from the edit history of Wikidata to create a model that can resolve constraint violations.
In contrast, our framework is general and does not assume any additional KB meta information or external data.

Regarding more general approaches, some aim at eliminating constraint violations.
For example, Chortis et al. \cite{chortis2015diagnosis,tonon2015fixing} defined and added new properties to avoid violating integrity constraints, 
while Melo \cite{de2013not} removed \textit{sameAs} links that lead to such violations.
These methods ensure KB consistency,
but they can neither correct the knowledge itself
nor deal with those wrong assertions that satisfy the constraints.
Lertvittayakumjorn et al. \cite{lertvittayakumjorn2017correcting} and Melo et al. \cite{melo2017approach} both aimed at correcting assertions by replacing the objects or subjects with correct entities.
The former found the substitute by either keyword matching or a simple graph structure metric (i.e., the number of commonly connected objects of two entities),
while the latter first retrieved candidate substitutes from the Wikipedia disambiguation page (which may not exist, especially for KBs that are not based on Wikipedia) and then ranked them by a lexical similarity score.
Both methods, however, only use simple graph structure or lexical similarity to identify the substitute, and ignore the linkage incompleteness of a KB.
In contrast, our framework utilizes state-of-the-art semantic embedding to model and exploit the local context within a sub-graph to predict assertion likelihood, 
and at the same time uses global property constraints to validate the substitution.

\section{Background}\label{sec:background}

\subsection{Knowledge Base}
In this study we consider a knowledge base (KB) that follows Semantic Web standards including RDF (Resource Description Framework), RDF Schema, OWL (Web Ontology Language)\footnote{There is a revision of the Web Ontology Language called OWL~2, for simplicity we also refer to this revision as OWL.} and the SPARQL Query Language \cite{domingue2011handbook}.
A KB is assumed to be composed of a TBox (terminology) and an ABox (assertions). 
The TBox usually defines classes (concepts), a class hierarchy (via \textit{rdfs:subClassOf}), properties (roles), and property domains and ranges.
It may also use a more expressive language such as OWL to express constraints such as class disjointness, property cardinality, property functionality and so on \cite{owl2}.

The ABox consists of a set of assertions (facts) describing concrete entities (individuals), each of which is represented by an Uniform Resource Identifier.
Each assertion is represented by an RDF triple $\langle s,p,o \rangle$,
where $s$ is an entity, $p$ is a property 
and $o$ is either an entity or a literal (i.e., a typed or untyped data value such as a string or integer).
$s$, $p$ and $o$ are known as the subject, predicate and object of the triple.
An entity can be an instance of
one or more classes, which is specified via triples using the \textit{rdf:type} property. Sometimes we will use \emph{class assertion} to refer to this latter kind of assertion and \emph{property assertion} to refer to assertions where $p$ is not a property from the reserved vocabulary or RDF, RDFS or OWL.

Such a KB can be accessed by SPARQL queries using a query engine that supports the relevant entailment regime (e.g., RDFS or OWL) \cite{glimm2012sparql}; such an engine can, e.g., infer $\left\langle e_0 \text{ \textit{rdf:type} } c_2 \right\rangle$,
given $\left\langle e_0 \text{ \textit{rdf:type} } c_1 \right\rangle$ and $\left\langle c_1 \text{ \textit{rdfs:subClassOf} } c_2 \right\rangle$.
In addition, large-scale KBs (aka knowledge graphs) often have a lookup service that enables users to directly access its entities by fuzzy matching; this is usually based on a lexical index that is built with entity labels (phrases defined by \textit{rdfs:label}) and sometimes entity anchor text (short descriptions).
DBpedia builds its lookup service\footnote{\url{https://wiki.dbpedia.org/lookup}} using the lexical index of Spotlight \cite{mendes2011dbpedia}, 
while entities of Wikidata can be retrieved, for example, via the API developed for OpenRefine and OpenTapioc \cite{delpeuch2019opentapioca}.

\subsection{Problem Statement}
In this study,
we focus on correcting ABox property assertions $\langle s,p,o \rangle$ where $o$ is a literal
(\textit{literal assertion}) or an entity (\textit{entity assertion}). Note that in the former
case correction may require more than simple canonicalization; e.g., the property assertion
$\langle$\textit{Sergio\_Ag\"{u}ero}, \textit{playsFor}, \textit{``Manchester United''}$\rangle$ should be
corrected to $\langle$\textit{Sergio\_Ag\"{u}ero}, \textit{playsFor}, \textit{Manchester\_City}$\rangle$.

Literal assertions can be identified by data type inference and regular expressions as in \cite{gunaratna2016gleaning},
while erroneous entity assertions can be detected either manually when the KB is applied 
in downstream applications or automatically by the methods discussed in Section~\ref{sec:rw_av}.
It is important to note that
if the KB is an OWL ontology, the set of \emph{object properties} (which connect two entities) and \emph{data properties} (which connect an entity to a literal) should be disjoint. 
In practice, however, KBs such as DBpedia often do not respect this constraint.

We assume that the input is a KB $\mathcal{K}$, and a set $\mathcal{E}$ of literal and/or entity assertions that have been identified as incorrect.
For each assertion $\langle s,p,o \rangle$ in $\mathcal{E}$, 
the proposed correction framework aims at either finding an entity $e$ from $\mathcal{K}$ as an object substitute, 
such that $e$ is semantically related to $o$ and the new triple $\langle s,p,e \rangle$ is true,
or reporting that there is no such an entity $e$ in $\mathcal{K}$.

\section{Methodology}\label{sec:method}

\subsection{Framework}
\begin{figure*}[h]
\centering
\includegraphics[scale=0.52]{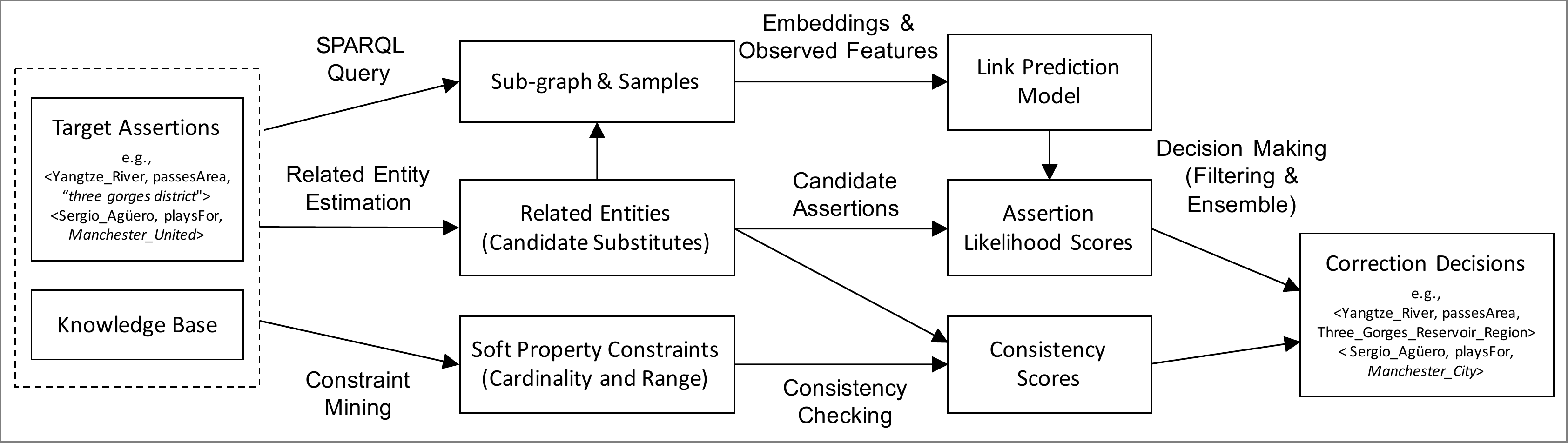}
\caption{
The Overall Framework for Assertion Correction}
\label{fig:framework}
\end{figure*}

As shown in Figure \ref{fig:framework}, the main components of our assertion correction framework consist of related entity estimation, link prediction, constraint-based validation and correction decision making.
Related entity estimation identifies those entities that are relevant to the object of the assertion.
Given a target assertion $t=\langle s,p,o \rangle$, its related entities, ranked by the relatedness, are denoted as $RE_t$.
They are called \textit{candidate substitutes} of the original object $o$, and the new assertions when $o$ is replaced are called \textit{candidate assertions}.
We adopt two techniques --- lexical matching and word embedding --- to measure relatedness and estimate $RE_t$.
Note that the aim of this step is to ensure high recall;
precision is subsequently taken care of via 
link prediction and constraint-based validation over the candidate assertions.

Link prediction estimates the likelihood of each candidate assertion.
For each entity $e_i$ in $RE_t$, it considers a target assertion $t=\langle s,p,o \rangle$ 
and outputs a score that measures the likelihood of $\langle s, p, e_i \rangle$.
To train such a link prediction model, a sub-graph that contains the context of the correction task (i.e., $\mathcal{E}$) is first extracted, 
with the related entities, involved properties and their neighbourhoods;
positive and negative assertions are then sampled for training.
State-of-the-art semantic embeddings (TransE \cite{bordes2013translating} and DistMult \cite{yang2015embedding}), as well as some widely used observed features (path and node)
are used to build the link prediction model.

Constraint-based validation 
checks whether a candidate assertion violates constraints on the cardinality or (hierarchical) range of the property,
and outputs a consistency score which measures its degree of consistency against such constraints. 
Such constraints can be effective in filtering out unlikely assertions, 
but modern KBs such as DBpedia and Wikidata often include only incomplete or weak constraints, 
or do not respect the given constraints as no global consistency checking is performed. 
Therefore, we do not assume that there are any property cardinality or range constraints in the KB TBox,\footnote{Any property range and cardinality constraints that are defined in the TBox, or that come from external knowledge, can be easily and directly injected into the framework.} 
but instead use mined constraints, 
each of which is associated with a supporting degree (probability).

Correction decision making combines the results of related entity estimation, link prediction and constraint-based validation; it first integrates the assertion likelihood scores and consistency scores,
and then filters out those candidate substitutes that have low scores.
Finally, it either reports that no suitable correction was found, or recommends the most likely correction.

\subsection{Related Entity Estimation}
For each target assertion $t = \langle s, p, o \rangle$ in $\mathcal{E}$, 
related entity estimation directly adopts $o$ as the input if $o$ is a literal, 
or extracts the label of $o$ if $o$ is an entity.
It returns a list $RE_t$ containing up to $k$ most related entities; i.e., $|RE_t| \le k$.
Our framework supports both a lexical matching based approach and a word embedding based approach; 
this allows us to compare the effectiveness of the two approaches on different KBs (see Section~\ref{sec:eva_ree}).

For those KBs with a lexical index, the lexical matching based approach can directly use a lookup service based on the index,
which often returns a set of related entities for a given phrase.
Direct lookup with the original phrase, however, often misses the correct entity, 
as the input phrase, either coming from the erroneous entity or the literal, is frequently noisy and ambiguous.
For example, the DBpedia Lookup service returns no entities with the input ``three gorges district'' which refers to the entity \textit{dbr:Three\_Gorges\_Reservoir\_Region}.
To improve recall, we retrieve a list of up to $k$ entities by repeating entity lookup using sub-phrases, starting with the longest sub-phrases and continuing with shorter and shorter sub-phrases until either $k$ entities have been retrieved or all sub-phrases have been used. 
The list of each lookup is ordered according to the relatedness (lexical similarity) to the original phrase, while all the lists are concatenated according to the above lookup order.
To extract the sub-phrases, we first tokenize the original phrase, remove the stop words
and then concatenate the tokens in their original order for sub-phrases of different lengths.
%
For those KBs without an existing lexical index, the lexical matching based approach adopts a string similarity score named Edit Distance \cite{navarro2001guided} in order to calculate the relatedness of an entity with its label.

The word embedding based approach calculates the similarity of $o$ against each entity in the KB, 
using vector representations of their labels (literals).
It \textit{(i)} tokenizes the phrase and removes the stop words, 
\textit{(ii)} represents each token by a vector with a word embedding model (e.g., \textit{Word2Vec} \cite{mikolov2013efficient}) that is trained using a large corpus, where tokens that are out of the model's vocabulary are ignored,
\textit{(iii)} calculates the average of the vectors of all the tokens,
which is a widely adopted strategy to embed a phrase,
and \textit{(iv)} computes the distance-based similarity score of the two vectors by e.g., the cosine similarity.

Compared with lexical matching, word embedding considers the semantics of a word, which assigns a high similarity score to two synonyms.
In the above lookup example, 
``district'' becomes noise as it is not included in the label of \textit{dbp:Three\_Gorges\_Reservoir\_Region},
but can still play an important role in the word embedding based approach due to the short word vector distance between ``district'' and ``region''.
However, in practice entity misuse is often not caused by semantic confusion, but by similarity of spelling and token composition, where the lexical similarity is high but the semantics might be quite different.
Moreover, lexical matching with a lexical index makes it easy to utilize multiple items of textual information, such as labels in multiple languages and anchor text, where different names of an entity are often included.

\subsection{Link Prediction}
Given related entities $RE_t$ of a target assertion $t=\langle s,p,o \rangle$,
link prediction is used to estimate a likelihood score for the candidate assertion $\langle s,p,e_i \rangle$,
for each entity $e_i$ in $RE_t$.
For efficiency in dealing with very large KBs, we first extract a multi-relational sub-graph for the context of the task,
and then train the link prediction model with a sampling method as well as with different observed features and  semantic embeddings.

\subsubsection{Sub-graph}
Given a KB $\mathcal{K}$ and a set of target assertions $\mathcal{E}$,
the sub-graph corresponding to $\mathcal{E}$ is a part of $\mathcal{K}$, 
denoted as $\mathcal{K_E} = \langle E, P, T \rangle$, where $E$ denotes entities, $P$ denotes object properties (relations) and $T$ denotes assertions (triples).
As shown in Algorithm~\ref{alg:sub_kb},
the sub-graph is calculated with three steps: 
\textit{(i)} extract the seeds --- entities and properties involved in the target assertions $\mathcal{E}$, 
as well as related entities of each assertion in $\mathcal{E}$;
\textit{(ii)} extract the neighbourhoods --- directly associated assertions of each of the seed properties and entities;
\textit{(iii)} re-calculate the properties and entities involved in the assertions.
Note that $\models$ means an assertion is either directly declared in the KB or can be inferred by the KB.
The statements with $\models$ can be implemented by SPARQL:
Line \ref{alg:line_models1} needs $\left| P \right|$ queries each of which retrieves the associated assertions of a given property, 
while Line \ref{alg:line_models2} needs $2 \times \left| E \right|$ queries 
each of which retrieves the associated assertions of a given subject or object.

\begin{algorithm}[h!]
\small
\KwIn{
\textit{(i)} The whole KB: $\mathcal{K}$,
\textit{(ii)} The set of target assertions: $\mathcal{E}$,
\textit{(iii)} The related entities of each target assertion: $RE_t$, $t\in \mathcal{E}$
} 
\KwResult{
The sub-graph: $\mathcal{K_E} = \langle E, P, T \rangle$
}
\Begin{
\emph{$\%$ Step 1: Extract the seeds} \\
$SE=\left\{s \vert \langle s,p,o \rangle \in \mathcal{E} \right\}$ \emph{$\%$ extract subject entities} \\
$P=\left\{p \vert \langle s,p,o \rangle \in \mathcal{E} \right\}$ \emph{$\%$ extract target properties} \\
$RE = \cup_{t \in \mathcal{E}}RE_t$ \emph{$\%$ The union of related entities} \\
$E=SE \cup RE$ \\
\ForEach{$\langle s,p,o \rangle$ in $\mathcal{E}$}{
\If{$o$ \text{ is an entity }}{
$E = E \cup \left\{ o \right\}$ \emph{$\%$ add object entity} \\}
}
\emph{$\%$ Step 2: Extract the neighbourhoods} \\
$T = \left\{ \langle s, p, o\rangle \vert p \in P  \land \mathcal{K} \models \langle s, p, o\rangle \right\}$ \label{alg:line_models1} \\
$T = T \cup \left\{ \langle s, p, o\rangle \vert o \in E \land s \in E \land \mathcal{K} \models \langle s,p,o \rangle  \right\}$\label{alg:line_models2} \\
\emph{$\%$ Step 3: Re-calculate entities and properties} \\
$E = E \cup \left\{s \vert \langle s,p,o\rangle \in T \right\} \cup \left\{o | \langle s,p,o\rangle \in T \right\}$ \\
$P = P \cup \left\{ p \vert \langle s,p,o \rangle \in T\right\}$ \\
\Return{$\mathcal{K_E} = \langle E, P, T \rangle$}
}
\caption{ 
\label{alg:sub_kb} 
\small{Sub-graph Extraction ($\mathcal{K}$, $\mathcal{E}$, $RE_t$)}
}

\end{algorithm}

\subsubsection{Sampling}
Positive and negative samples (assertions) are extracted from the sub-graph $\mathcal{K_E} = \langle E, P, T \rangle$.
The positive samples are composed of two parts:
$T_{pos} = T_{sp} \cup T_{pr}$,
where $T_{sp}$ refers to assertions whose subjects and properties are among $SE$ (i.e., those subject entities involved in $\mathcal{E}$) and $P$ respectively ($T_{sp} = \left\{ \langle s,p,o \rangle | s \in SE \land  p \in P \land  \langle s,p,o \rangle \in T \right\}$),
while $T_{pr}$ refers to those assertions whose objects and properties are among $RE$ (i.e., those related entities involved in $\mathcal{E}$) and $P$ respectively
($T_{pr} = \left\{ \langle s,p,o \rangle | p \in P \land o \in RE \land \langle s,p,o \rangle \in T \right\}$).
$T_{sp}$ and $T_{pr}$ are calculated by two steps: \textit{(i)} extract all the associated assertions of each property $p$ in $P$ from $T$;
\textit{(ii)} group these assertions according to $SE$ and $RE$.
Compared with an arbitrary assertion in $T$, 
the above samples are more relevant to the candidate assertions for prediction.
This can help release the domain adaption problem --- the data distribution gap between the training and predicting assertions.

The negative samples are composed of two parts as well: $T_{neg} = \tilde{T}_{sp} \cup \tilde{T}_{pr}$,
where $\tilde{T}_{sp}$ is constructed according to $T_{sp}$ by replacing the object with a random entity in $E$,
while $\tilde{T}_{pr}$ are constructed according to $T_{pr}$ by replacing the subject with a random entity in $E$.
Take $T_{sp}$ as an example, for each of its assertion $\langle s,p,o \rangle$,
an entity $\tilde{e}$ is randomly selected from $E$ for a synthetic assertion  $\tilde{t} = \langle s,p,\tilde{e}\rangle$ such that $\mathcal{K} \nvDash \tilde{t}$,
where $\nvDash$ represents that an assertion is neither declared by the KB nor can be inferred,
and $\tilde{t}$ is added to $\tilde{T}_{sp}$.
In implementation, we can get $\mathcal{K} \nvDash \tilde{t}$ if $\tilde{t} \notin T$, as $T$ is extracted from the KB with inference.
$T_{pr}$ is constructed similarly.
Note that the size of $T_{pos}$ and $T_{neg}$ is balanced.

\subsubsection{Observed Features}
We extract two kinds of observed features --- the path feature and the node feature.
The former represents potential relations that can connect the subject and object,
while the latter represents the likelihood of the subject being the head of the property, and the likelihood of the object being the tail of the property.
For the path feature, we limit the path depth to two, for reducing computation time and feature size, both of which are exponential w.r.t. the depth.
In fact, it has been shown that paths of depth one are already quite effective. 
They outperform the state-of-the-art KB embedding methods like DistMult, TransE and TransH, together with the node feature on some benchmarks \cite{toutanova2015observed}.
Meanwhile the predictive information of a path will vanish as its depth increases. 

In calculation,
we first extract paths of depth one: $FP^1_{so}$ and $FP^1_{os}$, 
where $FP^1_{so}$ represents properties from $s$ to $o$ (i.e., $\left\{p_0 | \langle s,p_0,o\rangle \in T \right\}$),
while  $FP^1_{os}$ represents properties from $o$ to $s$ (i.e., $\left\{p_0 | \langle o,p_0,s\rangle  \in T\right\}$).
Next we calculate paths of depth two (ordered property pairs) in two directions as well: 
$FP^2_{so} = \left\{(p_1,p_2) | \langle s,p_1,e\rangle \in T \land \langle e,p_2,o\rangle \in T \right\}$, 
$FP^2_{os} = \left\{(p_1,p_2) | \langle o,p_1,e\rangle \in T \land \langle e, p_2,s\rangle \in T \right\}$.
Finally we merge these paths: $FP = FP^1_{so} \cup FP^1_{os} \cup FP^2_{so} \cup FP^2_{os}$, 
and encode them into a multi-hot vector as the path feature, denoted as $f^p$.
Briefly we collect all the unique paths from the training assertions as a candidate set,
where one path corresponds to one slot in encoding.
When an assertion is encoded into a vector, 
a slot of the vector is set to $1$ if the slot's corresponding path is among the assertion's paths and $0$ otherwise.

The node feature includes two binary variables: $f^n = \left[ v_s, v_o\right]$, where $v_s$ denotes the likelihood of the subject while $v_o$ denotes the likelihood of the object.
Namely, $v_s = 1$ if there exists some entity $o^{\prime}$ such that $\langle s, p, o^{\prime}\rangle \in T$ and $v_s = 0$ otherwise.
$v_o = 1$ if there exists some entity $s^{\prime}$ such that $\langle s^{\prime}, p, o\rangle \in T$ and $v_o=0$ otherwise. 



Finally we calculate $f^p$ and $f^n$, and concatenate them for each sample in $T_{pos} \cup T_{neg}$, 
and train a link prediction model with a basic supervised classifier named Multiple Layer Perception (MLP):
\begin{equation}
    \mathcal{M}_{pn}  \xleftarrow[T_{pos} \cup T_{neg}]{\text{classifier e.g., MLP}} \left[f^{p}, f^n \right].
\end{equation}

We also adopt the path-based latent feature learned by the state-of-the-art algorithm RDF2Vec \cite{ristoski2016rdf2vec}, as a baseline.
RDF2Vec first extracts potential outstretched paths of an entity by e.g., graph walks, 
and then learns embeddings of the entities through the neural language model \textit{Word2Vec}.
In training,
we encode the subject and object of an assertion by their RDF2Vec embeddings,
encode its property by a one-hot vector,
concatenate the three vectors,
and use the same classifier MLP.
The trained model is denoted as $\mathcal{M}_{r2v}$.

\subsubsection{Semantic Embeddings}
A number of semantic embedding algorithms have been proposed to learn the vector representation of KB properties and entities.
One common way is to define a scoring function to model the truth of an assertion,
and use an overall loss for learning.
We adopt two state-of-the-art algorithms --- TransE \cite{bordes2013translating} and DistMult \cite{yang2015embedding}.
Both are simple but have been shown to be competitive or outperform more complex alternatives \cite{yang2015embedding,bordes2013translating,kadlec2017knowledge}. 
For high efficiency, 
we learn the embeddings from the sub-graph.

TransE tries to learn a vector representation space 
such that $o$ is a nearest neighbor of $s + p$ if an assertion $\langle s,p,o \rangle$ holds, 
and $o$ is far away from $s + p$ otherwise.
$+$ denotes the vector add operation.
To this end, the score function of $t =\langle s,p,o \rangle$, denoted as $g(t)$,
is defined as $d(\bm{e}_s + \bm{e}_p, \bm{e}_o)$, where $d$ is a dissimilarity (distance) measure such as $L_2$ norm, 
while $\bm{e}_s$, $\bm{e}_p$ and $\bm{e}_o$ are embeddings of $s$, $p$ and $o$ respectively.
The embeddings have the same dimension that is configured, and are initialized by one-hot encoding. 
In learning, a batch stochastic gradient descent algorithm is used to minimize the the following margin-based ranking loss:
\begin{equation}\label{eq:TransE_loss}
   L = \sum_{t \in T, t \rightarrow \tilde{t}}  \left[ \gamma + g(t) - g(\tilde{t}) \right]_{+}
\end{equation}
where $\gamma > 0$ is a hyper parameter, $\left[ \cdot \right]_+$ denotes extracting the positive part, and $\tilde{t}$ represents a negative assertion of $t$, generated by randomly replacing the subject or object with an entity in $E$. 

DistMult is a special form of the bilinear model where the non-diagonal entries in the relation matrices are assumed to be zero.
The score function of an assertion $\langle s,p,o \rangle$ is defined as $\bm{e}_p^T(\bm{e}_s \circ \bm{e}_o)$, where $\circ$ denotes the operation of pairwise multiplication.
As TransE, the embeddings are initialized by one-hot encoding, with the a dimension configured.
A similar margin-based ranking loss as \eqref{eq:TransE_loss} is used for training with batched stochastic gradient descent.

In prediction, the likelihood score of an assertion can be calculated with the corresponding scoring function and the embeddings of its subject, property and object.
We denote the link prediction model by TransE and DistMult as $\mathcal{M}_{te}$ and $\mathcal{M}_{dm}$ respectively. 
  

\subsection{Constraint-based Validation}
We first mine two kinds of soft constraints 
--- property cardinality and hierarchical property range from the KB,
and then use a consistency checking algorithm to validate those candidate assertions.

\subsubsection{Property Cardinality}
Given a property $p$,
its soft cardinality is represented by a probability distribution $D^p_{car}(k) \in \left[0, 1\right]$, 
where $k \ge 1$ is an integer that denotes the cardinality.
It is calculated as follows: 
\textit{(i)} get all the property assertions whose property is $p$, denoted as $T(p)$, 
and all the involved subjects, denoted as $S(p)$,
\textit{(ii)} count the number of the object entities associated with each subject $s$ in $S(p)$ and $p$:
$ON(s,p) = \left| \left\{ o | \langle s,p,o \rangle \in T(p) \right\} \right| $,
\textit{(iii)} find out the maximum object number: $ON_{max}^p = max \left\{ ON(s,p) | s \in S(p) \right\}$,
and \textit{(iv)} calculate the property cardinality distribution as:
\begin{equation}
D^p_{car}(k) = \frac{\left| \left\{ s \in S(p) | ON(s,p)=k \right\} \right|}{\left| S(p) \right|}, k = 1,...,ON_{max}^p, 
\end{equation}
where $\left| \cdot \right|$ denotes the size of a set. 
Specially $ON_{max}^p = 0$ if $T(p)$ is empty.
$D_{car}^{p}(k > n)$ is short for  $\sum_{i=n+1}^{ON_{max}^p}D_{car}^p(k=i)$,
denoting the probability that the cardinality is larger than $n$.
In implementation, $T(p)$ can be accessed by one time SPARQL query, 
while the remaining computation has linear time complexity w.r.t. $\left| T(p) \right|$.

The probability of cardinality $k$ is equal to the ratio of the subjects that are associated with $k$ different entity objects.
For example, considering a property \textit{hasParent} that is associated with $10$ different subjects (persons) in the KB,
if one of them has one object (parent) and the remaining have two objects (parents), then the cardinality distribution is: $D_{car}(k=1) = \sfrac{1}{10}$ and $D_{car}(k=2) = \sfrac{9}{10}$.
Note that although such constraints follow Closed Word Assumption and Unique Name Assumption,
they behave well in our method.
On the one hand, probabilities are estimated to represent the supporting degree of a constraint by the ABox.
One the other hand, they are used in an approximate model to validate candidate assertions 
instead of as new and totally true knowledge for KB TBox extension.

\subsubsection{Hierarchical Property Range}
Given a property $p$, its range constraint consists of \textit{(i)} \textit{specific range} which includes the most specific classes of its associated objects, denoted as $C_{sp}^p$,
and \textit{(ii)} \textit{general range} which includes ancestors of these most specific classes, denoted as $C_{ge}^p$, with top classes such as \textit{owl:Thing} being excluded.
A most specific class of an entity refers to one of the most fine grained classes 
that the entity is an instance of according to class assertions in the KB. Note
that there could be multiple such classes as the entity could be asserted to be
an instance of multiple classes for which there is no sub-class relationship.
General classes of an entity are those that subsume one or more of the specific classes
as specified in the KB via \textit{rdfs:subClassOf} assertions.

Each range class $c$ in $C_{sp}^p$ ($C_{ge}^p$ resp.) has a probability in $[0,1]$ that 
represents its supporting degree by the KB, denoted as $D_{sp}^p(c)$ ($D_{ge}^p(c)$ resp.).
 $C_{sp}^p$, $C_{ge}^p$ and the supporting degrees are calculated by the following steps:
\textit{(i)} get all the object entities that are associated with $p$, denoted as $OE(p)$;
\textit{(ii)} infer the specific and general classes of each entity $oe$ in $OE(p)$, denoted as $C_{sp}(p, oe)$ and $C_{ge}(p, oe)$ respectively, 
and at the same time collect $C_{sp}^p$ as $\cup_{oe \in OE(p)}C_{sp}(p,oe)$ and $C_{ge}^p$  as $\cup_{oe \in OE(p)}C_{ge}(p,oe)$;
\textit{(iii)} compute the supporting degrees:
\begin{equation}
\begin{cases}
    D_{sp}^p(c) = \frac{\left| \left\{ oe | oe \in OE(p), c \in C_{sp}(p, oe) \right\} \right|}{\left| OE(p) \right|}, c \in C_{sp}^p, \\
    D_{ge}^p(c) = \frac{\left| \left\{ oe | oe \in OE(p), c \in C_{ge}(p, oe) \right\} \right|}{\left| OE(p) \right|},c \in C_{ge}^p. \\
    \end{cases}
\end{equation}
The degree of each range class is the ratio of the objects that are instances of the class, as either directly declared in the ABox or inferred by \textit{rdfs:subClassOf}.
The implementation needs one time SPARQL query to get $OE(p)$, and 
$\left| OE(p) \right|$ times SPARQL queries to get the specific and ancestor classes. 
The remaining computation has linear time complexity w.r.t.  $\left| OE(p) \right|$.
As property cardinality, 
the range cardinality is also used for approximating the likelihood of candidate assertions, 
using a consistency checking algorithm introduced bellow.

\subsubsection{Consistency Checking}
As shown in Algorithm \ref{alg:constraint_checking},
constraint checking acts as a model, to estimate the consistency of an assertion against soft constraints of hierarchical property range and cardinality.
Given a candidate assertion $t=\langle s,p,e \rangle$,
the algorithm first checks the property cardinality,
with a parameter named maximum cardinality exceeding rate $\sigma \in (0,1]$.
Line \ref{alg_l:count} counts the number of entity objects that are associated with $s$ and $p$ in the KB, 
assuming that the correction is made (i.e., $t$ has been added into the KB).
Note that $1 \le n \le ON_{max}^p + 1$.
Line \ref{alg_1:exceed_r} calculates its exceeding rate $r$ w.r.t. $ON_{max}^p$, where $r \in (-\infty,1]$.
In Line \ref{alg_l:zero_condition}, $ON_{max}^p = 0$ indicates that $p$ is highly likely to used as a data property in the KB.
This is common in correcting literal assertions: one example is the property \textit{hasName} whose objects are phrases of entity mentions but should not be replaced by entities.
In this case, it is more reasonable to report that the object substitute does not exist, and thus the algorithm sets the cardinality score $y_{car}$ to $0$.

Another condition of setting $y_{car}$ to $0$ is $r \ge \sigma$. 
Specially, when $\sigma$ is set to $1.0$,
$r \ge \sigma$ (i.e.,  $ON_{max}^p = 1$ , $n = 2$) means that 
$p$ is a object property with functionality in the KB but the correction violates this constraint.
Note that $n$ can exceed $ON_{max}^p$ by a small degree which happens when $ON_{max}^p$ is large.
For example, when $\sigma$ is set to $0.5$, $r = 0.25$ (i.e., $ON_{max}^p=4$ and $n=5$) is allowed.
Line \ref{alg_l:fun_condition} to \ref{alg_l:non_fun_score} calculate the property cardinality score $y_{car}$
as the probability of being a functional property ($n=1$),
or as the probability of being a none-functional property ($n>1$).
Specially, we punish the score when $n > ON_{max}^p$ (i.e., $r > 0$) by multiplying it with a degrading factor $1-r$: the higher exceeding rate, the more it degrades. 

Line \ref{alg_l:classes} to \ref{alg_l:range_end} calculate the property range score $y_{ran}$, 
by combing the specific range score $y_{ran,c}$ and the general range score $y_{ran,g}$ with their importance weights $\omega_c$ and $\omega_g$. 
Usually we make the specific range more important by setting $\omega_c$ and $\omega_g$ to e.g., $0.8$ and $0.2$ respectively.
Line \ref{alg_l:range_scores} computes $y_{ran,c}$ and $y_{ran,g}$: 
the score is higher if more classes of the objects are among the range classes,
and these classes have higher range degrees.
For example, considering the property \textit{bornIn} with the following range cardinality: $C_{sp}^p = \left\{ City, Town, Place\right\}$, $C_{ge}^p = \left\{ Place \right\}$,
$D_{sp}^p(City) = 0.5$, $D_{sp}^p(Town) = 0.4$, $D_{sp}^p(Place) = 0.05$ and $D_{ge}^p(Place) = 0.95$,
we will have 
\textit{(i)} $y_{ran,c} = 1 - (1-0.5)(1-0.05) = 0.525$ and $y_{ran,g} = 0.95$ if $C(e) = \left\{ City, Place\right\}$,
\textit{(ii)} $y_{ran,c} = 0.05$ and $y_{ran,g} = 0.95$ if $C(e) = \left\{ Village, Place \right\}$,
and \textit{(iii)} $y_{ran,c} = 0$ and $y_{ran,g} = 0$ if $C(e) = \left\{ Professor, Person \right\}$.
The order of the consistency degree against the property range is: $\left\{ City, Place \right\} > \left\{ Village, Place \right\} > \left\{ Professor, Person \right\}$.

The algorithm finally returns the property cardinality score $y_{car}$ and the property range score $y_{ran}$.
The former model is denoted as  $\mathcal{M}_{car}$ while the letter is denoted as $\mathcal{M}_{ran}$.
According to some empirical analysis, we can multiply or average the two scores, as the final model of consistency checking, denoted as $\mathcal{M}_{car+ran}$.

\begin{algorithm}[h!]
\small
\KwIn{
\textit{(i)} A candidate assertion: $t = \langle s,p,e \rangle$,
\textit{(ii)} property cardinality constraint: $\langle D_{car}^p, ON_{max}^p \rangle$,
\textit{(iii)} the maximum cardinality exceeding rate: $\sigma \in (0,1]$,
\textit{(iv)} hierarchical property range constraint: $\langle D_{sp}^p, D_{ge}^p, C_{sp}^p, C_{ge}^p \rangle$,
\textit{(v)} weights of the specific range and general range: $\langle \omega_c, \omega_g \rangle$
} 
\KwResult{
$y_{car}$: score that $t$ is consistent with the property cardinality; $y_{ran}$: score that $t$ is consistent with the property range
}
\Begin{
\emph{$\%$ count the number of object entities} \label{alg_l:car_start}\\
$n = \left| \left\{ o | \mathcal{K} \models \langle s, p, o\rangle \text{, $o$ is entity} \right\} \cup \left\{ e \right\}\right|$;\label{alg_l:count} \\ 
$r = \frac{(n-ON_{max}^p)}{ON_{max}^p}$; \emph{$\%$ calculate the exceeding rate} \label{alg_1:exceed_r}\\
\emph{$\%$ no object entities are associated with $p$ in the KB, or the cardinality exceeds the maximum by a specific rate} \\
\eIf{$ON_{max}^p = 0 \parallel r \ge \sigma $\label{alg_l:zero_condition}}{
$y_{car} = 0$; \\
}{
\eIf{$n = 1$\label{alg_l:fun_condition}}{
\emph{$\%$ probability as a functional property} \\
$y_{car} = D_{car}^p(k=1)$; \label{alg_l:fun_score}\\
}{
\emph{$\%$ probability as a none-functional property} \\
$y_{car} =
\begin{cases}
D_{car}^p(k>1), & \mbox{if } r \le 0 \\
(1-r) \cdot D_{car}^p(k>1), & \mbox{else}
\end{cases}
$\label{alg_l:non_fun_score}\\
}
}
$C(e) = \left\{ c | \mathcal{K} \models \langle e, \textit{rdf:type}, c\rangle\right\}$; \emph{$\%$ get the object's classes} \label{alg_l:classes}\\
\emph{$\%$ calculate the constraint score of specific and general ranges\label{alg_l:range_start}} \\
$
\begin{cases}
y_{ran,c} = 1 - \prod_{c\in C_{sp}^p \cap C(e)}(1-D_{ran}^p(c)), \\
y_{ran,g} = 1 - \prod_{c\in C_{ge}^p \cap C(e)}(1-D_{ran}^p(c));
\end{cases}$\label{alg_l:range_scores}\\
\emph{$\%$ calculate the overall range constraint score\label{alg_l:overall_score_comment}} \\
$y_{ran} = \omega_{c} \cdot y_{ran,c} + \omega_{g} \cdot y_{ran,g}$;\label{alg_l:range_end} \\
\Return{$y_{car}$, $y_{ran}$}
}
\caption{ 
\label{alg:constraint_checking} 
\small{Consistency Checking ($\mathcal{M}_{ran}$, $\mathcal{M}_{car}$)
}
}

\end{algorithm}

\subsection{Correction Decision Making}
Given a target assertion $t$ in $\mathcal{E}$, and its top-$k$ related entities $RE_t$, 
for each entity $e_i$ in $RE_t$,
the correction framework \textit{(i)} calculates the assertion likelihood score $y^l_{i}$ with a link prediction model ($\mathcal{M}_{pn}$, $\mathcal{M}_{te}$ or $\mathcal{M}_{dm}$),
and the consistency score $y^c_{i}$  with $\mathcal{M}_{car}$, $\mathcal{M}_{ran}$ or $\mathcal{M}_{car+ran}$;
\textit{(ii)} separately normalizes $y^l_{i}$ and $y^c_{i}$ into $\left[0,1 \right]$  according to all the predictions by the corresponding model for $\mathcal{E}$;
\textit{(iii)} ensembles the two scores by simple averaging: 
$y_i = \sfrac{(y^l_{i} + y^c_{i})}{2}$;
\textit{(iv)} filters out $e_i$ from $RE_t$ if $y_{i} < \tau$.
Note $e_i$ is always kept if $t$ is a literal assertion and its literal is exactly equal to the label of $e_i$.
The related entities after filtering keep their original order in $RE_t$, and are denoted as $RE_t^{\prime}$.
$\tau$ is a parameter in $\left[0,1 \right]$ that needs to be adjusted with a developing data set.
It eventually returns none, which means there is no entity in the KB that can replace the object of $t$, if $RE_t^{\prime}$ is empty,
and the top-$1$ entity in $RE_t^{\prime}$ as the object substitute otherwise.
The ensemble of the link prediction score and constraint-based validation score is not a must.
Either of them can make a positive impact independently, while their ensemble can make the performance higher in most cases, as evaluated in Section \ref{sec:eva_overall}.

\section{Evaluation}\label{sec:eva}
\subsection{Experiment Settings}
\subsubsection{Data}
In our experiment, we correct assertions in DBpedia \cite{auer2007dbpedia} 
and in an enterprise medical KB whose TBox is defined by clinic experts and ABox is extracted from medical articles (text) by some open information extraction tools (cf. more details in \cite{niklaus2018survey}).
DBpedia is accessed via its official Lookup service, SPARQL Endpoint\footnote{\url{http://dbpedia.org/sparql}} and entity label dump (for related entity estimation with \textit{Word2Vec}).
The medical KB contains knowledge about disease, medicine, treatment, symptoms, foods and so on, 
with around $800$ thousand entities, $7$ properties, $48$ classes, $4$ million property assertions.
The data are representative of two common situations: errors of DBpedia are mostly inherited from the source while errors of the medical KB are mostly introduced by extraction.

Regarding DBpedia, we reuse the real world literal set proposed by \cite{chen2019canonicalizing,gunaratna2016gleaning}.
As our task is not typing the literal, but substituting it with an entity,
literals containing multiple entity mentions are removed,
while properties with insufficient literal objects are complemented with more literals from DBpedia.
We annotate each assertion with a ground truth (GT), which is either a correct 
replacement entity from DBpedia (i.e., \textit{Entity GT}) or none (i.e., \textit{Empty GT}). 
Ground truths are carefully checked using DBpedia, Wikipedia, and multiple external resources.
Regarding the medical KB, we use a set of assertions with erroneous entity objects
that have been discovered and collected during deployment of the KB in enterprise products;
the GT annotations have been added with the help of clinical experts. 
For convenience, we call the above two target assertion sets DBP-Lit and MED-Ent respectively.\footnote{DBP-Lit data and its experiment codes: \url{https://github.com/ChenJiaoyan/KG_Curation}}
More details are shown in Table \ref{res:statistics}.

\begin{table}[h!]
\small{
\centering
\renewcommand{\arraystretch}{1.3}
\begin{tabular}[t]{c|c|c|c}\hline
& Assertions (with Entity GT) \# & Properties \#& Subjects \#  \\ \hline
DBP-Lit & $725$ ($499$) & $127$ & $668$    \\ \hline
MED-Ent &$272$ ($225$)  &  $7$ &$200$    \\ \hline
\end{tabular}
\caption{\small
Some statistics of DBP-Lit and MED-Ent.
}\label{res:statistics}
}
\end{table}

\subsubsection{Settings}
In the evaluation, we first analyze related entity estimation (Section \ref{sec:eva_ree}) and link prediction (Section \ref{sec:eva_lp}) independently.
For related entity estimation, we report the recall of Entity GTs of different methods with varying top-$k$ values,
based on which a suitable method and a $k$ value are selected for the framework.
For link prediction, we compare the performance of different semantic embeddings and observed features,
using those target assertions whose  Entity GTs are recalled in related entity estimation.
The related entities of a target assertion are first ranked according to the predicted score,
and then standard metrics including Hits@$1$, Hits@$5$ and MRR (Mean Reciprocal Rank)\footnote{\url{https://en.wikipedia.org/wiki/Mean_reciprocal_rank}} are calculated.

Next we evaluate the overall results of the assertion correction framework (Section \ref{sec:eva_overall}),
where the baselines are compared with,
and the impact of link prediction and constraint-based validation is analyzed.
Three metrics are adopted: \textit{(i) Correction Rate} which is the ratio of the target assertions that are corrected with right substitutes, among all the target assertions with Entity GTs; 
\textit{(ii) Empty Rate} which is the ratio of the target assertions that are corrected with none, among all the target assertions with Empty GTs;
\textit{(iii) Accuracy} which is the ratio of the truly corrected target assertions by either substitutes or none, among all the target assertions.
Note that accuracy is an overall metric considering both correction rate and empty rate. 
Either high (low resp.) correction rate or empty rate can lead to high (low resp.) accuracy.
With the overall results, we finally analyze the constraint-based validation with more details.

The reported results are based on the following setting (unless otherwise specified).
In related entity estimation, \textit{Word2Vec} \cite{mikolov2013efficient} trained using the Wikipedia article dump in June 2018
is used for word embedding.
In link prediction,
the margin hyper parameter $\gamma$ is set by linear increasing w.r.t. the training step, 
the embedding size of both entities and properties is set to $100$,
and the other training hyper parameters such as the number of epochs and MLP hidden layer size are set such that the highest MRR is achieved on an evaluation sample set.
Regarding the baseline RDF2Vec, pre-trained versions of DBpedia entities with different settings by Mannheim University\footnote{\url{https://bit.ly/2M4TQOg}} are tested, and the results with the best MRR are reported.
In constraint-based validation, $\sigma$, $\omega_c$ and $\omega_g$ are set to $1.0$, $0.8$ and $0.2$ respectively, according to the algorithm insight. 
Some other reasonable settings explored can achieve similar results. 
The embeddings are trained by GeForce GTX 1080 Ti with OpenKE \cite{han2018openke},  
while the remaining is computed by Intel(R) Xeon(R) CPU E5-2670 @2.60GHz and 32G RAM.

\subsection{Related Entity Estimation}\label{sec:eva_ree}
\begin{figure}[h]
\centering
\includegraphics[scale=0.4]{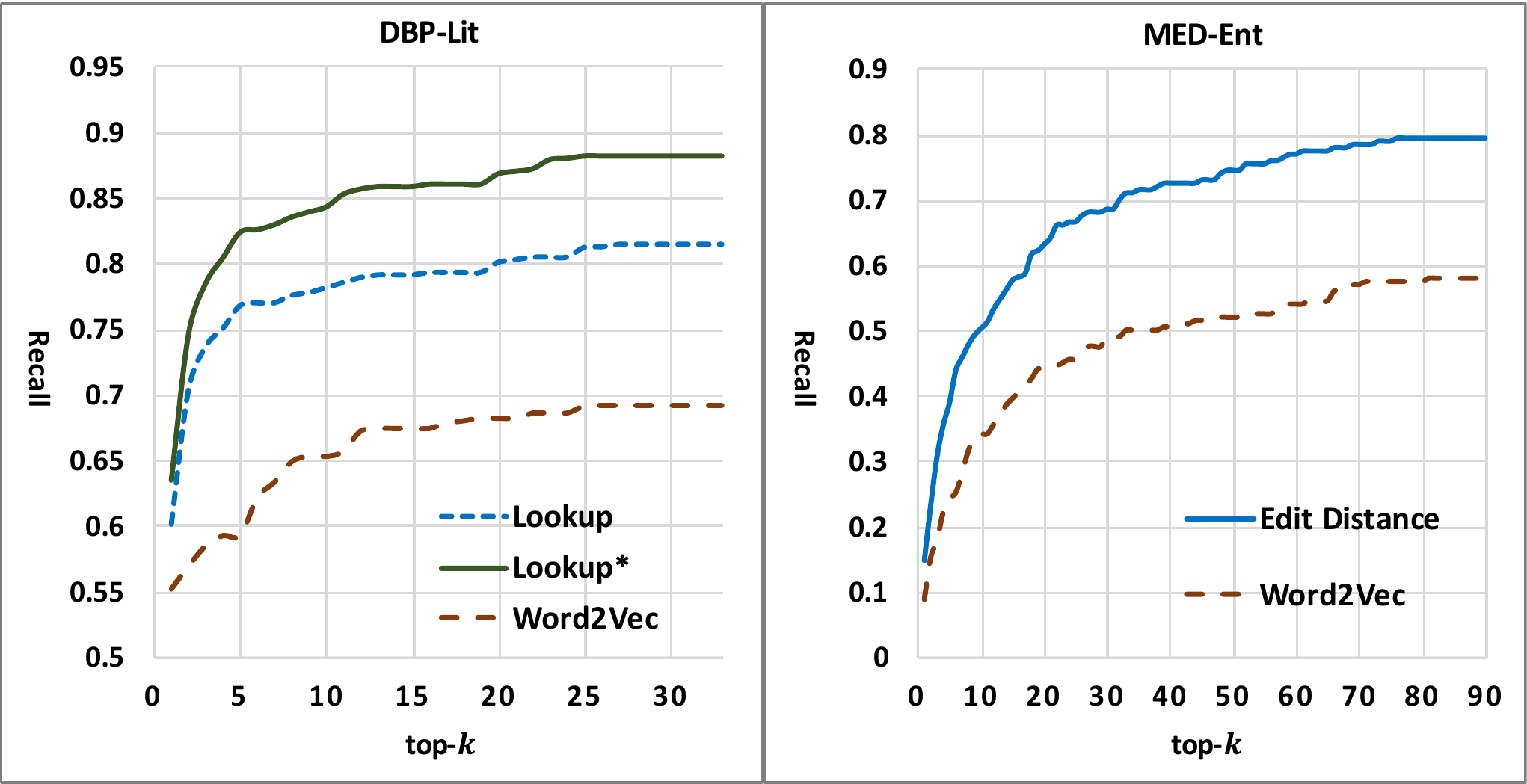}
\caption{\small
The recall of Entity GTs by top-$k$ related entities}
\label{fig:REE}
\end{figure}

We compare different methods and settings used in related entity estimation with the results presented in Figure \ref{fig:REE},
where the recall of Entity GTs by top-$k$ related entities are presented.
First, we find that the lexical matching based methods (Lookup, $\text{Lookup}^{*}$ and Edit Distance) have much higher recall than \textit{Word2Vec}, on both DBP-Lit and MED-Ent.
The reason for DBP-Lit may lie in the Lookup service provided by DBpedia,
which takes not only the entity label but also the anchor text into consideration.
The latter provides more semantics, some of which, such as different names and background description,
is very helpful for recalling the right entity.
The reason for MED-Ent, according to some empirical analysis, 
is that the erroneous objects are often caused by lexical confusion,
such as misspelling and misusing of an entity with similar tokens.
Second, our Lookup solution with sub-phrases, i.e., $\text{Lookup}^{*}$, as expected, outperforms the original Lookup.
For example, when both curves are stable, their recalls are around $0.88$ and $0.81$ respectively, 

The target of related entity estimation in our framework is to have a high recall with a $k$ value that is not too large (so as to avoid additional noise and limit the size of the sub-graph for efficiency). 
In real application, the method and $k$ value can be set by analyzing the recall.
According to the absolute value and the trend of the recall curves in Figure \ref{fig:REE}, 
our framework uses $\text{Lookup}^{*}$ with $k=30$ for DBP-Lit, and Edit Distance with $k=76$ for MED-Ent.

\subsection{Link Prediction}\label{sec:eva_lp}
\subsubsection{Impact of Models}
The results of different link prediction methods are shown in Table \ref{res:lp}, where the sub-graph is used for training.
The baseline Random means randomly ranking the related entities, 
while AttBiRNN refers to the attentive bidirectional Recurrent Neural Networks that utilize the labels of the subject, property and object.
AttBiRNN was used in \cite{chen2019canonicalizing} for literal object typing, with good performance achieved.
First of all, the results verify that either latent semantic embeddings (TransE and DistMult) or observed features with  Multiple Layer Perception are effective for both DBP-Lit and MED-Ent: MRR, Hits@1 and Hits@5 are all dramatically improved in comparison with Random and AttBiRNN. 

We also find that concatenating the node feature and path feature (Node+Path) achieves higher performance than the node feature and the path feature alone, as well as the baseline RDF2Vec which is based on graph walks.
For DBP-Lit, the outperformance over RDF2Vec is $39.9\%$, $44.1\%$ and $45.5\%$ for MRR, Hits@1 and Hits@5 respectively.

Meanwhile, Node+Path performs better than TransE and DistMult for DBP-Lit,
while for MED-Ent, TransE and DistMult outperforms Node+Path.
For example, considering the metric of MRR, 
Node+Path is $71.3\%$ higher than DistMult for DBP-Lit, but DistMul is $108.8\%$ higher than Node+Path for MED-Ent.
One potential reason is the difference in the number of properties and sparsity of the two sub-graphs.
DBP-Lit has $127$ properties in its target assertions and $1958$ properties in its sub-graph;
while MED-Ent has $7$ properties in its target assertions and $19$ properties in its sub-graph.
The small number of properties for MED-Ent leads to quite poor path feature, which is verified by its independent performance (e.g., the MRR is only $0.09$).
In the sub-graph of DBP-Lit,
the average number of connected entities per property (i.e., density) is $150.7$, while in the sub-graph of MED-Ent, it is $2739.0$. 
Moreover, a larger ratio of properties to entities also leads to richer path features.
According to these results, we use Node+Path for DBP-Lit and DistMult for MED-Ent in our correction framework.

\begin{table}[h!]
\small{
\centering
\renewcommand{\arraystretch}{1.3}
\begin{tabular}[t]{c||p{0.76cm}<{\centering}|p{0.77cm}<{\centering}|p{0.77cm}<{\centering}||p{0.76cm}<{\centering}|p{0.77cm}<{\centering}|p{0.77cm}<{\centering}}\hline
\multirow{2}{*}{Methods} & \multicolumn{3}{|c||}{DBP-Lit}  & \multicolumn{3}{|c}{MED-Ent}   \\ \cline{2-7}
 & MRR & Hits@1 & Hits@5 & MRR & Hits@1 & Hits@5 \\ \hline
Random &$0.199$ &$0.100$  &$0.275$  &$0.027$ &$0.013$ &$0.066$   \\ \hline
AttBiRNN &$0.251$ &$0.126$ &$0.348$ &$0.255$ &$0.111$ &$0.414$   \\ \hline
TransE &$0.342$ & $0.173$ &$0.518$  &$0.744$ &$0.652$ &$\bm{0.862}$   \\ 
DistMult &$0.424$ &$0.300$ &$0.536$  &$\bm{0.752}$ &$\bm{0.694}$ &$0.806$   \\ \hline
RDF2Vec &$0.419$ &$0.320$  &$0.492$  &--- &--- &---   \\ 
Node &$0.495$ &$0.379$ &$0.604$  &$0.338$ &$0.171$ &$0.514$   \\ 
Path &$0.473$ &$0.356$ &$0.578$  &$0.090$ &$0.028$ &$0.133$   \\ 
Node+Path &$\bm{0.586}$ &$\bm{0.461}$ &$\bm{0.716}$ &$0.360$ &$0.200$ &$0.525$   \\ \hline
\end{tabular}
\caption{\small
Link prediction results based on the sub-graph.
}\label{res:lp}
}
\end{table}

\subsubsection{Impact of The Sub-graph}
We further analyze the impact of sub-graph usage in training the link prediction model.
The results of some of the methods that can be run over the whole KB with limited time are shown in Table \ref{res:whole_kb},
where Node+Path (DBP-Lit) uses features extracted from the whole KB but samples from the sub-graph.
One the one hand, 
in comparison with Node+Path trained purely with the sub-graph, 
Node+Path with global features actually performs worse.
As all the directly connected properties and entities of each subject entity, related entity and target property are included in the sub-graph, 
using the sub-graph makes no difference for node features and path features of depth one.
Thus the above result is mainly due to the fact that path features of depth two actually makes limited contribution in this link prediction context.
This is reasonable as they are weak, duplicated or even noisy in comparison with node features and path features of depth one.
One the other hand, learning the semantic embeddings with the sub-graph  has positive impact on TransE and negative impact on DistMult for MED-Ent.
However the impact in both cases is quite limited.
Considering that the sub-graph has only $36.9\%$ ($11.2\%$ resp.) of the entities (assertions resp.) of the whole medical KB, 
which reduces the training time of DistMult embeddings from $46.7$ minutes to $19.0$ minutes,
the above performance drop can be accepted.

\begin{table}[h!]
\small{
\centering
\renewcommand{\arraystretch}{1.2}
\begin{tabular}[t]{p{2.44cm}<{\centering}|c|c|c}\hline
Cases & MRR  & Hits@1  & Hits@5   \\ \hline
TransE (MED-Ent) &$0.713$ (-$.031$) &$0.608$ (-$.044$) & $0.834$ (-$.028$)  \\ \hline
DistMult (MED-Ent) &$0.766$ (+$.014$) &$0.721$ (+$.027$) &$0.822$ (+$.016$)   \\ \hline
Node+Path (DBP-Lit) &$0.504$ (-$.082$) &$0.384$ (-$.077$) &$0.611$ (-$.105$)   \\ \hline
\end{tabular}
\caption{\small
Link prediction results based on the whole KB, and their outperformance (gap) over those based on the sub-graph
}\label{res:whole_kb}
}
\end{table}

\subsection{Overall Results}\label{sec:eva_overall}
\begin{figure}
\begin{subfigure}
  \centering
  \includegraphics[width=.99\linewidth]{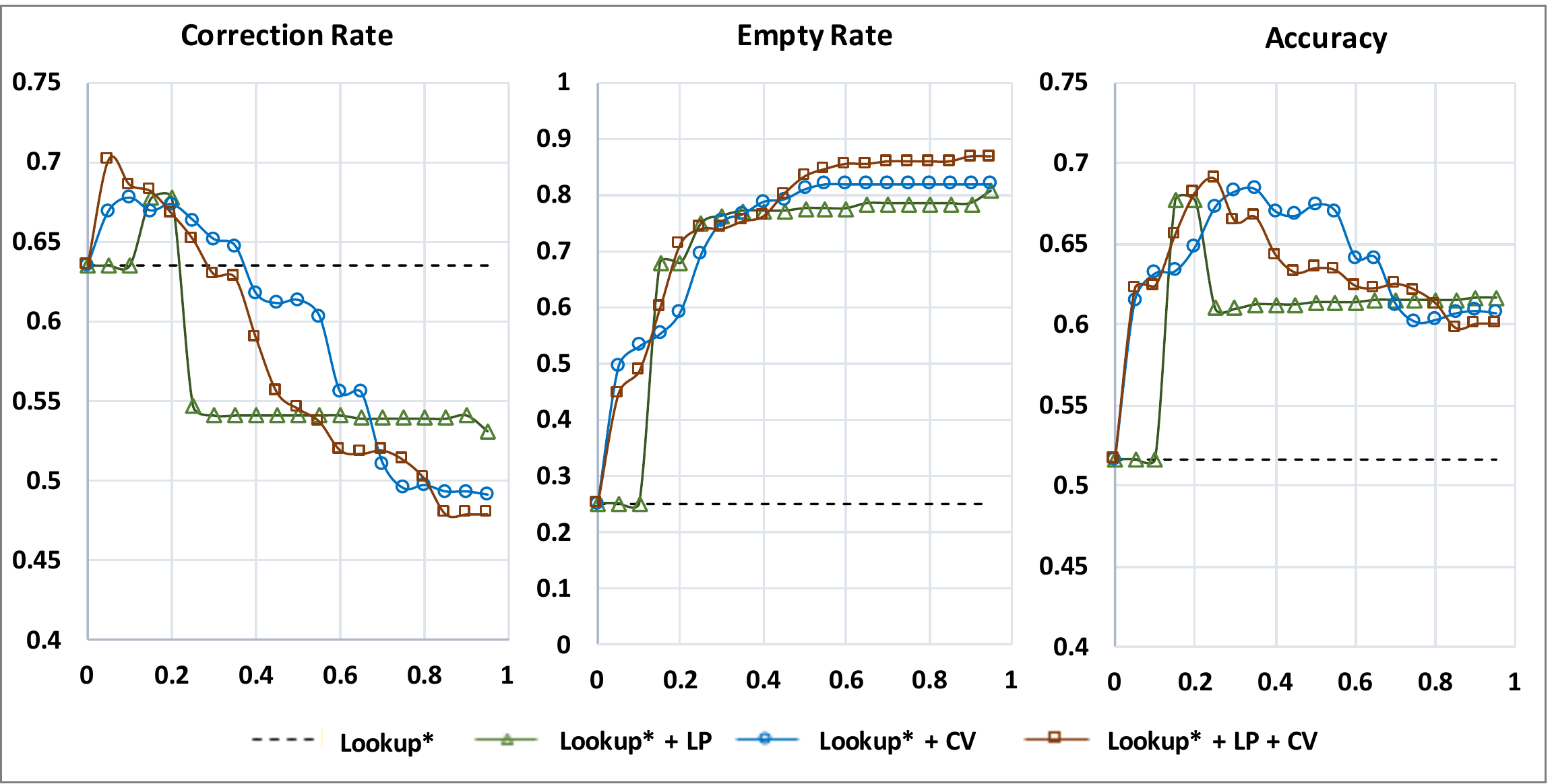}  
\end{subfigure}
\begin{subfigure}
  \centering
  \includegraphics[width=.995\linewidth]{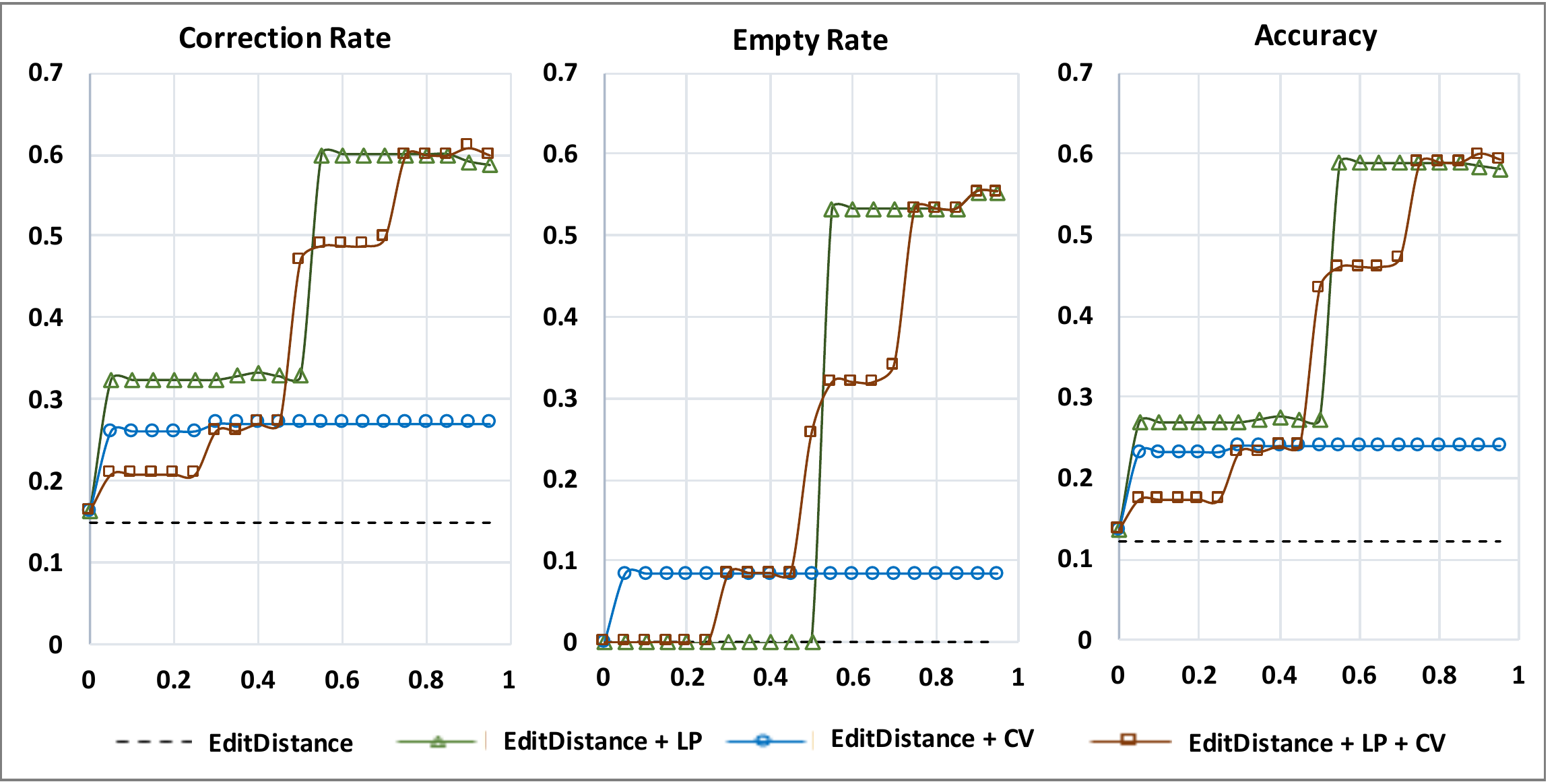}  
\end{subfigure}
\caption{\small
Overall results of the correction framework for DBP-Lite [Above] and MED-Ent [Below]. 
+ LP and + CV represent filtering with link prediction and constraint-based validation respectively, with the filtering threshold $\bm{\tau}$ ranging from $\bm{0}$ to $\bm{1}$ with a step of $\bm{0.05}$.
}
\label{fig:overall}
\end{figure}

Figure \ref{fig:overall} presents the correction rate, empty rate and accuracy  of our assertion correction framework with a ranging filtering threshold $\tau$.
Note that lexical matching without any filtering is similar to the existing method discussed in related work \cite{lertvittayakumjorn2017correcting}. 
On the one hand, we find that filtering  with either link prediction (LP) or constraint-based validation (CV) 
can improve the correction rate when $\tau$ is set to a suitable range. 
This is because those candidate substitutes that are lexically similar to the erroneous object but lead to unlikely assertions are filtered out,
while those that are not so lexically similar but lead to true assertions are ranked higher.
As the empty rate is definitely increased after filtering (e.g., improved from $0.252$ to $0.867$ by $\text{Lookup}^{*}$ + LP + CV for DBP-Lit),
the accuracy for both DBP-Lit and MED-Ent is improved in the whole range of $\tau$.
On the other hand, we find that averaging the scores from link prediction and constraint-based validation is effective.
It leads to both higher correction rate and accuracy than either of them for some values of $\tau$, such as $[0.05,0.1 ]$ for DBP-Lit and $[0.85, 0.95]$ for MED-Ent.

Table \ref{res:optimum} presents the optimum correction rate and accuracy for several settings.
Note that they are achieved using a suitable $\tau$ setting; in real applications this can be determined using an evaluation data set.
Based on these results, we make the following observations.
First, the optimum results are consistent with our above conclusions regarding the positive impact of link prediction, constraint-based validation and their ensemble.
For example, the optimum accuracy of DBP-Lit is improved by $32.6\%$ using constraint-based validation in comparison with the original related entities using lexical matching.
The correction rate of MED-Ent provides another example:
REE + LP + CV is $1.5\%$ higher than REE + LP,
and $121.4\%$ higher than REE + CV.

Second, lexical matching using either Lookup (for DBP-Lit) or Edit Distance (for MED-Ent) has a much higher correction rate and accuracy than  \textit{Word2Vec},
while our Lookup with sub-phrases ($\text{Lookup}^{*}$) has even higher correction rate than the original Lookup of DBpedia. 
These overall results verify the recall analysis on related entity estimation in Section~\ref{sec:eva_ree}. 
Meanwhile, we find that the overall results on observed features ($\mathcal{M}_{np}$) and latent semantic embeddings ($\mathcal{M}_{dm}$ and $\mathcal{M}_{te}$)
are also consistent with the link prediction analysis in Section~\ref{sec:eva_lp}: $\mathcal{M}_{np}$ has a better filtering effect than $\mathcal{M}_{dm}$ and $\mathcal{M}_{te}$ for DBP-Lit, but worse filtering effect for MED-Ent.

\begin{table}[h!]
\small{
\centering
\renewcommand{\arraystretch}{1.3}
\begin{tabular}[t]{c||p{0.8cm}<{\centering}|p{0.78cm}<{\centering}||p{0.8cm}<{\centering}|p{0.78cm}<{\centering}}\hline
\multirow{2}{*}{Methods} & \multicolumn{2}{|c||}{DBP-Lit}  & \multicolumn{2}{|c}{MED-Ent}   \\ \cline{2-5}
 & C-Rate & Acc & C-Rate & Acc  \\ \hline
Lexical Matching  & $0.597$  & $0.611$ &$0.149$  &$0.123$    \\ 
$\text{Lookup}^*$ &$0.635$  &$0.516$  &--- &---   \\ 
\textit{Word2Vec} &$0.553$  &$0.410$  &$0.089$ &$0.076$   \\ \hline\hline
REE + LP ($\mathcal{M}_{np}$) &$0.677$  &$0.677$  &$0.360$ &$0.327$   \\ 
REE +  LP ($\mathcal{M}_{dm}$) &$0.635$ &$0.628$  & $0.600$&$0.588$   \\ \hline\hline
REE +  CV ($\mathcal{M}_{ran}$) &$0.671$  &$0.668$  &$0.271$ &$0.239$   \\ 
REE +  CV ($\mathcal{M}_{car}$) & $0.639$&$0.622$  &$0.164$ &$0.147$   \\ 
REE + CV ($\mathcal{M}_{ran+car}$) & $0.677$  &$0.684$  &$0.271$ &$0.246$   \\ \hline\hline
REE + LP + CV &$\bm{0.701}$  & $\bm{0.690}$ &$\bm{0.609}$ &$\bm{0.599}$   \\ \hline
\end{tabular}
\caption{\small
Optimum correction rate (C-Rate) and accuracy (Acc).
For Lexical Matching, DBP-Lit uses the original DBpedia Lookup while MED-Ent uses Edit Distance.
REE denotes Related Entity Estimation: DBP-Lit uses $\text{Lookup}^*$ while MED-Ent uses Edit Distance.
}\label{res:optimum}
}
\end{table}

\subsection{Constraint-based Validation}

Besides the positive impact on the overall results mentioned above,
we get several more detailed observations about constraint-based validation from Table~\ref{res:optimum}.
On the one hand, the property range constraint plays a more important role than the property cardinality constraint,
while their combination 
is more robust than either of them, as expected.
Considering the assertion set of MED-Ent, 
filtering by $\mathcal{M}_{ran}$, for example, leads to $62.6\%$ higher accuracy than filtering by $\mathcal{M}_{car}$,
while filtering by $\mathcal{M}_{ran+car}$ has $2.9\%$ higher accuracy and equal correction rate in comparison with $\mathcal{M}_{ran}$.

On the other hand, we find constraint-based validation performs well for DBP-Lit, with higher accuracy and equal correction rate in comparison with link prediction,
but performs much worse for MED-Ent. 
This is mainly due to the gap between the semantics of the two target assertion sets and their corresponding KBs:
\textit{(i)} the mined property ranges of DBP-Lit include $404$ hierarchical classes,
while those of MED-DB have only $8$ classes in total and these classes have no hierarchy;
\textit{(ii)} $23$ out of $127$ target properties in DBP-Lit have pure functionality (i.e., $D^p_{car}(k=1) = 1.0$) which plays a key role in the consistency checking algorithm, 
while none of the target properties of MED-Ent has such pure functionality.
The second characteristic is also a potential reason why filtering by constraint-based validation with property cardinality only achieves a very limited improvement over Edit Distance for MED-Ent as shown in Table \ref{res:optimum}.

We additionally present some examples of the mined soft property constraints in Table~\ref{res:examples}.
Most of them are consistent with our common sense understanding of the properties, although some noise is evident (e.g., the range Person and Agent of \textit{dbp:homeTown}), most likely caused by erroneous property and class assertions.

\begin{table}[h!]
\small{
\centering
\renewcommand{\arraystretch}{1.2}
\begin{tabular}{m{1.75cm}<{\centering}|m{1.25cm}<{\centering}|m{1.9cm}<{\centering}|m{2.1cm}<{\centering}}\hline
Property & Cardinality  & Specific Range  & General Range   \\ \hline
\textit{dbp:homeTown} & $1: 0.415$ $2:0.422$ $3:0.163$ &Location: $0.801$, City: $0.280$, Country: $0.268$, Person: $0.228$, ... &PopulatedPlace: $0.821$, Place: $0.821$, Settlement: $0.299$, Agent: $0.259$, ...   \\ \hline
\textit{dbp:finalteam} &$1: 1.000$ &BaseballTeam: $0.493$, SportsTeam: $0.154$, SoccerClub: $0.015$, ... &Agent: $0.688$, Organization: $0.665$, SportsTeam: $0.510$, ...   \\ \hline
\end{tabular}
\caption{\small
Soft constraints of two property examples
}\label{res:examples}
}
\end{table}

\section{Discussion and Outlook}\label{sec:dis}
In this paper we present a study of assertion correction,
an important problem for KB curation, but one that has rarely been studied.
We have proposed a general correction framework,
which does not rely on any KB meta data or external information,
and which exploits both both deep learning and consistency reasoning to correct erroneous objects and informally annotated literals (entity mentions).
The framework and the adopted techniques have been evaluated by correcting assertions in two different KBs: DBpedia with cross-domain knowledge and an enterprise KB from the medical domain. 
We discuss below several more observations of the study as well as possible directions for the future work.

\textbf{Entity relatedness.}
Our method follows the principle of correcting the object by a related entity 
rather than an arbitrary entity that leads to a correct assertion.
Relatedness can be due to either lexical or semantic similarity.
Currently, the recall for DBP-Lit and MED-Ent is $0.882$ and $0.797$ respectively, 
which is promising, but still leaves space for further improvement. 
One extension for higher recall but limited noise and sub-graph size is incorporating external resources like Wikipedia disambiguation pages or domain knowledge about the errors.

\textbf{KB variation.}
Although both constraint-based validation  and  link prediction improve overall performance, 
their impact varies from DBpedia to the medical KB.
The effectiveness of constraint-based validation depends on the richness of the KB schema,
such as property functionality, 
the complexity of property range, etc.
The more complex the schema is, the better performance constraint-based validation achieves.
The impact of link prediction is more complicated: the path and node features perform better on DBpedia which has many more properties than the medical KB,
while the semantic embeddings by DistMult and TransE are more suitable for the medical KB which has less properties but higher density. 
Integrating link prediction and constraint-based validation, even with simple score averaging, can improve performance, 
but further study is needed for a better integration method that is adapted to the structure of the given KB.

\textbf{Property constraints.}
On the one hand, the evaluation indicates that the mined property constraints are effective for assertion validation
and can be independently used in other contexts like online KB editing.
On the other hand, unlike the link prediction model, 
the constraints as well as the consistency checking algorithm are interpretable.
One benefit is that explicitly defined or external TBox constraints can easily be injected into our framework by overwriting or revising the constraints.
For example, the mined specific range Person: $0.228$ in Table \ref{res:examples}, which is inappropriate for the property \textit{dbp:homeTown}, can be directly removed.

\section*{Acknowledgements}
The work is supported by
the AIDA project, The Alan Turing Institute under the EPSRC
grant EP/N510129/1, 
the SIRIUS Centre for Scalable Data Access (Research Council of Norway, project 237889),
the Royal Society,
EPSRC projects DBOnto, $\text{MaSI}^{\text{3}}$~and~$\text{ED}^{\text{3}}$. 
A part of the data and computation resource as well as Xi Chen's contribution are supported by Jarvis Lab Tencent.

\bibliographystyle{ACM-Reference-Format}
\balance
\bibliography{bibliography}


\begin{thebibliography}{47}


\ifx \showCODEN    \undefined \def \showCODEN     #1{\unskip}     \fi
\ifx \showDOI      \undefined \def \showDOI       #1{#1}\fi
\ifx \showISBNx    \undefined \def \showISBNx     #1{\unskip}     \fi
\ifx \showISBNxiii \undefined \def \showISBNxiii  #1{\unskip}     \fi
\ifx \showISSN     \undefined \def \showISSN      #1{\unskip}     \fi
\ifx \showLCCN     \undefined \def \showLCCN      #1{\unskip}     \fi
\ifx \shownote     \undefined \def \shownote      #1{#1}          \fi
\ifx \showarticletitle \undefined \def \showarticletitle #1{#1}   \fi
\ifx \showURL      \undefined \def \showURL       {\relax}        \fi
\providecommand\bibfield[2]{#2}
\providecommand\bibinfo[2]{#2}
\providecommand\natexlab[1]{#1}
\providecommand\showeprint[2][]{arXiv:#2}

\bibitem[\protect\citeauthoryear{Arndt, De~Meester, Dimou, Verborgh, and
  Mannens}{Arndt et~al\mbox{.}}{2017}]%
        {arndt2017using}
\bibfield{author}{\bibinfo{person}{D{\"o}rthe Arndt}, \bibinfo{person}{Ben
  De~Meester}, \bibinfo{person}{Anastasia Dimou}, \bibinfo{person}{Ruben
  Verborgh}, {and} \bibinfo{person}{Erik Mannens}.}
  \bibinfo{year}{2017}\natexlab{}.
\newblock \showarticletitle{Using Rule-based Reasoning for RDF Validation}. In
  \bibinfo{booktitle}{\emph{International Joint Conference on Rules and
  Reasoning}}. Springer, \bibinfo{pages}{22--36}.
\newblock


\bibitem[\protect\citeauthoryear{Auer, Bizer, Kobilarov, Lehmann, Cyganiak, and
  Ives}{Auer et~al\mbox{.}}{2007}]%
        {auer2007dbpedia}
\bibfield{author}{\bibinfo{person}{S{\"o}ren Auer}, \bibinfo{person}{Christian
  Bizer}, \bibinfo{person}{Georgi Kobilarov}, \bibinfo{person}{Jens Lehmann},
  \bibinfo{person}{Richard Cyganiak}, {and} \bibinfo{person}{Zachary Ives}.}
  \bibinfo{year}{2007}\natexlab{}.
\newblock \showarticletitle{{DBpedia: A Nucleus for A Web of Open Data}}.
\newblock In \bibinfo{booktitle}{\emph{The Semantic Web}}.
  \bibinfo{publisher}{Springer}, \bibinfo{pages}{722--735}.
\newblock


\bibitem[\protect\citeauthoryear{Bordes, Usunier, Garcia-Duran, Weston, and
  Yakhnenko}{Bordes et~al\mbox{.}}{2013}]%
        {bordes2013translating}
\bibfield{author}{\bibinfo{person}{Antoine Bordes}, \bibinfo{person}{Nicolas
  Usunier}, \bibinfo{person}{Alberto Garcia-Duran}, \bibinfo{person}{Jason
  Weston}, {and} \bibinfo{person}{Oksana Yakhnenko}.}
  \bibinfo{year}{2013}\natexlab{}.
\newblock \showarticletitle{Translating Embeddings for Modeling
  Multi-relational Data}. In \bibinfo{booktitle}{\emph{Advances in Neural
  Information Processing Systems}}. \bibinfo{pages}{2787--2795}.
\newblock


\bibitem[\protect\citeauthoryear{Chen, Jim{\'{e}}nez{-}Ruiz, and Horrocks}{Chen
  et~al\mbox{.}}{2019}]%
        {chen2019canonicalizing}
\bibfield{author}{\bibinfo{person}{Jiaoyan Chen}, \bibinfo{person}{Ernesto
  Jim{\'{e}}nez{-}Ruiz}, {and} \bibinfo{person}{Ian Horrocks}.}
  \bibinfo{year}{2019}\natexlab{}.
\newblock \showarticletitle{Canonicalizing Knowledge Base Literals}. In
  \bibinfo{booktitle}{\emph{International Semantic Web Conference}}.
\newblock


\bibitem[\protect\citeauthoryear{Chortis and Flouris}{Chortis and
  Flouris}{2015}]%
        {chortis2015diagnosis}
\bibfield{author}{\bibinfo{person}{Michalis Chortis} {and}
  \bibinfo{person}{Giorgos Flouris}.} \bibinfo{year}{2015}\natexlab{}.
\newblock \showarticletitle{{A Diagnosis and Repair Framework for DL-LiteA
  KBs}}. In \bibinfo{booktitle}{\emph{European Semantic Web Conference}}.
  Springer, \bibinfo{pages}{199--214}.
\newblock


\bibitem[\protect\citeauthoryear{De~Melo}{De~Melo}{2013}]%
        {de2013not}
\bibfield{author}{\bibinfo{person}{Gerard De~Melo}.}
  \bibinfo{year}{2013}\natexlab{}.
\newblock \showarticletitle{Not quite the same: Identity constraints for the
  web of linked data}. In \bibinfo{booktitle}{\emph{Twenty-Seventh AAAI
  Conference on Artificial Intelligence}}.
\newblock


\bibitem[\protect\citeauthoryear{Debattista, Lange, and Auer}{Debattista
  et~al\mbox{.}}{2016}]%
        {debattista2016preliminary}
\bibfield{author}{\bibinfo{person}{Jeremy Debattista},
  \bibinfo{person}{Christoph Lange}, {and} \bibinfo{person}{S{\"o}ren Auer}.}
  \bibinfo{year}{2016}\natexlab{}.
\newblock \showarticletitle{A Preliminary Investigation towards Improving
  Linked Data Quality Using Distance-based Outlier Detection}. In
  \bibinfo{booktitle}{\emph{Joint International Semantic Technology
  Conference}}. Springer, \bibinfo{pages}{116--124}.
\newblock


\bibitem[\protect\citeauthoryear{Delpeuch}{Delpeuch}{2019}]%
        {delpeuch2019opentapioca}
\bibfield{author}{\bibinfo{person}{Antonin Delpeuch}.}
  \bibinfo{year}{2019}\natexlab{}.
\newblock \showarticletitle{OpenTapioca: Lightweight Entity Linking for
  Wikidata}.
\newblock \bibinfo{journal}{\emph{arXiv preprint arXiv:1904.09131}}
  (\bibinfo{year}{2019}).
\newblock


\bibitem[\protect\citeauthoryear{Dimou, Kontokostas, Freudenberg, Verborgh,
  Lehmann, Mannens, Hellmann, and Van~de Walle}{Dimou et~al\mbox{.}}{2015}]%
        {dimou2015assessing}
\bibfield{author}{\bibinfo{person}{Anastasia Dimou}, \bibinfo{person}{Dimitris
  Kontokostas}, \bibinfo{person}{Markus Freudenberg}, \bibinfo{person}{Ruben
  Verborgh}, \bibinfo{person}{Jens Lehmann}, \bibinfo{person}{Erik Mannens},
  \bibinfo{person}{Sebastian Hellmann}, {and} \bibinfo{person}{Rik Van~de
  Walle}.} \bibinfo{year}{2015}\natexlab{}.
\newblock \showarticletitle{Assessing and Refining Mappings to RDF to Improve
  Dataset Quality}. In \bibinfo{booktitle}{\emph{International Semantic Web
  Conference}}. Springer, \bibinfo{pages}{133--149}.
\newblock


\bibitem[\protect\citeauthoryear{Domingue, Fensel, and Hendler}{Domingue
  et~al\mbox{.}}{2011}]%
        {domingue2011handbook}
\bibfield{author}{\bibinfo{person}{John Domingue}, \bibinfo{person}{Dieter
  Fensel}, {and} \bibinfo{person}{James~A Hendler}.}
  \bibinfo{year}{2011}\natexlab{}.
\newblock \bibinfo{booktitle}{\emph{Handbook of Semantic Web Technologies}}.
\newblock \bibinfo{publisher}{Springer Science \& Business Media}.
\newblock


\bibitem[\protect\citeauthoryear{F{\"a}rber, Bartscherer, Menne, and
  Rettinger}{F{\"a}rber et~al\mbox{.}}{2018}]%
        {farber2018linked}
\bibfield{author}{\bibinfo{person}{Michael F{\"a}rber},
  \bibinfo{person}{Frederic Bartscherer}, \bibinfo{person}{Carsten Menne},
  {and} \bibinfo{person}{Achim Rettinger}.} \bibinfo{year}{2018}\natexlab{}.
\newblock \showarticletitle{Linked Data Quality of DBpedia, Freebase, Opencyc,
  Wikidata, and Yago}.
\newblock \bibinfo{journal}{\emph{Semantic Web}} \bibinfo{volume}{9},
  \bibinfo{number}{1} (\bibinfo{year}{2018}), \bibinfo{pages}{77--129}.
\newblock


\bibitem[\protect\citeauthoryear{Gal{\'a}rraga, Heitz, Murphy, and
  Suchanek}{Gal{\'a}rraga et~al\mbox{.}}{2014}]%
        {galarraga2014canonicalizing}
\bibfield{author}{\bibinfo{person}{Luis Gal{\'a}rraga}, \bibinfo{person}{Geremy
  Heitz}, \bibinfo{person}{Kevin Murphy}, {and} \bibinfo{person}{Fabian~M
  Suchanek}.} \bibinfo{year}{2014}\natexlab{}.
\newblock \showarticletitle{Canonicalizing Open Knowledge Bases}. In
  \bibinfo{booktitle}{\emph{Proceedings of the 23rd ACM International
  Conference on Information and Knowledge Management}}. ACM,
  \bibinfo{pages}{1679--1688}.
\newblock


\bibitem[\protect\citeauthoryear{Glimm and Ogbuji}{Glimm and Ogbuji}{2013}]%
        {glimm2012sparql}
\bibfield{author}{\bibinfo{person}{Birte Glimm} {and} \bibinfo{person}{Chimezie
  Ogbuji}.} \bibinfo{year}{2013}\natexlab{}.
\newblock \showarticletitle{SPARQL 1.1 Entailment Regimes}.
\newblock \bibinfo{journal}{\emph{W3C Recommendation}} (\bibinfo{year}{2013}).
\newblock


\bibitem[\protect\citeauthoryear{Grau, Horrocks, Motik, Parsia,
  Patel{-}Schneider, and Sattler}{Grau et~al\mbox{.}}{2008}]%
        {owl2}
\bibfield{author}{\bibinfo{person}{Bernardo~Cuenca Grau}, \bibinfo{person}{Ian
  Horrocks}, \bibinfo{person}{Boris Motik}, \bibinfo{person}{Bijan Parsia},
  \bibinfo{person}{Peter~F. Patel{-}Schneider}, {and} \bibinfo{person}{Ulrike
  Sattler}.} \bibinfo{year}{2008}\natexlab{}.
\newblock \showarticletitle{{OWL} 2: The Next Step for {OWL}}.
\newblock \bibinfo{journal}{\emph{Web Semantics: Science, Services and Agents
  on the World Wide Web}} \bibinfo{volume}{6}, \bibinfo{number}{4}
  (\bibinfo{year}{2008}), \bibinfo{pages}{309--322}.
\newblock


\bibitem[\protect\citeauthoryear{Gunaratna, Thirunarayan, Sheth, and
  Cheng}{Gunaratna et~al\mbox{.}}{2016}]%
        {gunaratna2016gleaning}
\bibfield{author}{\bibinfo{person}{Kalpa Gunaratna},
  \bibinfo{person}{Krishnaprasad Thirunarayan}, \bibinfo{person}{Amit Sheth},
  {and} \bibinfo{person}{Gong Cheng}.} \bibinfo{year}{2016}\natexlab{}.
\newblock \showarticletitle{Gleaning Types for Literals in RDF Triples with
  Application to Entity Summarization}. In \bibinfo{booktitle}{\emph{European
  Semantic Web Conference}}. \bibinfo{pages}{85--100}.
\newblock


\bibitem[\protect\citeauthoryear{Han, Cao, Lv, Lin, Liu, Sun, and Li}{Han
  et~al\mbox{.}}{2018}]%
        {han2018openke}
\bibfield{author}{\bibinfo{person}{Xu Han}, \bibinfo{person}{Shulin Cao},
  \bibinfo{person}{Xin Lv}, \bibinfo{person}{Yankai Lin},
  \bibinfo{person}{Zhiyuan Liu}, \bibinfo{person}{Maosong Sun}, {and}
  \bibinfo{person}{Juanzi Li}.} \bibinfo{year}{2018}\natexlab{}.
\newblock \showarticletitle{OpenKE: An Open Toolkit for Knowledge Embedding}.
  In \bibinfo{booktitle}{\emph{Proceedings of the 2018 Conference on Empirical
  Methods in Natural Language Processing: System Demonstrations}}.
  \bibinfo{pages}{139--144}.
\newblock


\bibitem[\protect\citeauthoryear{Kadlec, Bajgar, and Kleindienst}{Kadlec
  et~al\mbox{.}}{2017}]%
        {kadlec2017knowledge}
\bibfield{author}{\bibinfo{person}{Rudolf Kadlec}, \bibinfo{person}{Ondrej
  Bajgar}, {and} \bibinfo{person}{Jan Kleindienst}.}
  \bibinfo{year}{2017}\natexlab{}.
\newblock \showarticletitle{Knowledge Base Completion: Baselines Strike Back}.
  In \bibinfo{booktitle}{\emph{Proceedings of the 2nd Workshop on
  Representation Learning for NLP}}. \bibinfo{pages}{69--74}.
\newblock


\bibitem[\protect\citeauthoryear{Knublauch and Kontokostas}{Knublauch and
  Kontokostas}{2017}]%
        {knublauch2017shapes}
\bibfield{author}{\bibinfo{person}{Holger Knublauch} {and}
  \bibinfo{person}{Dimitris Kontokostas}.} \bibinfo{year}{2017}\natexlab{}.
\newblock \showarticletitle{Shapes Constraint Language (SHACL)}.
\newblock \bibinfo{journal}{\emph{W3C Recommendation}}  \bibinfo{volume}{20}
  (\bibinfo{year}{2017}).
\newblock


\bibitem[\protect\citeauthoryear{Kontokostas, Westphal, Auer, Hellmann,
  Lehmann, Cornelissen, and Zaveri}{Kontokostas et~al\mbox{.}}{2014}]%
        {kontokostas2014test}
\bibfield{author}{\bibinfo{person}{Dimitris Kontokostas},
  \bibinfo{person}{Patrick Westphal}, \bibinfo{person}{S{\"o}ren Auer},
  \bibinfo{person}{Sebastian Hellmann}, \bibinfo{person}{Jens Lehmann},
  \bibinfo{person}{Roland Cornelissen}, {and} \bibinfo{person}{Amrapali
  Zaveri}.} \bibinfo{year}{2014}\natexlab{}.
\newblock \showarticletitle{Test-driven Evaluation of Linked Data Quality}. In
  \bibinfo{booktitle}{\emph{Proceedings of the 23rd International Conference on
  World Wide Web}}. ACM, \bibinfo{pages}{747--758}.
\newblock


\bibitem[\protect\citeauthoryear{Lertvittayakumjorn, Kertkeidkachorn, and
  Ichise}{Lertvittayakumjorn et~al\mbox{.}}{2017}]%
        {lertvittayakumjorn2017correcting}
\bibfield{author}{\bibinfo{person}{Piyawat Lertvittayakumjorn},
  \bibinfo{person}{Natthawut Kertkeidkachorn}, {and} \bibinfo{person}{Ryutaro
  Ichise}.} \bibinfo{year}{2017}\natexlab{}.
\newblock \showarticletitle{Correcting Range Violation Errors in DBpedia.}. In
  \bibinfo{booktitle}{\emph{International Semantic Web Conference (Posters,
  Demos \& Industry Tracks)}}.
\newblock


\bibitem[\protect\citeauthoryear{Melo and Paulheim}{Melo and Paulheim}{2017a}]%
        {melo2017approach}
\bibfield{author}{\bibinfo{person}{Andr{\'e} Melo} {and} \bibinfo{person}{Heiko
  Paulheim}.} \bibinfo{year}{2017}\natexlab{a}.
\newblock \showarticletitle{An Approach to Correction of Erroneous Links in
  Knowledge Graphs}. In \bibinfo{booktitle}{\emph{CEUR Workshop Proceedings}},
  Vol.~\bibinfo{volume}{2065}. RWTH, \bibinfo{pages}{54--57}.
\newblock


\bibitem[\protect\citeauthoryear{Melo and Paulheim}{Melo and Paulheim}{2017b}]%
        {melo2017detection}
\bibfield{author}{\bibinfo{person}{Andr{\'e} Melo} {and} \bibinfo{person}{Heiko
  Paulheim}.} \bibinfo{year}{2017}\natexlab{b}.
\newblock \showarticletitle{Detection of Relation Assertion Errors in Knowledge
  Graphs}. In \bibinfo{booktitle}{\emph{Proceedings of the Knowledge Capture
  Conference}}. ACM, \bibinfo{pages}{22}.
\newblock


\bibitem[\protect\citeauthoryear{Mendes, Jakob, Garc{\'\i}a-Silva, and
  Bizer}{Mendes et~al\mbox{.}}{2011}]%
        {mendes2011dbpedia}
\bibfield{author}{\bibinfo{person}{Pablo~N Mendes}, \bibinfo{person}{Max
  Jakob}, \bibinfo{person}{Andr{\'e}s Garc{\'\i}a-Silva}, {and}
  \bibinfo{person}{Christian Bizer}.} \bibinfo{year}{2011}\natexlab{}.
\newblock \showarticletitle{DBpedia Spotlight: Shedding Light on the Web of
  Documents}. In \bibinfo{booktitle}{\emph{Proceedings of the 7th International
  Conference on Semantic Systems}}. ACM, \bibinfo{pages}{1--8}.
\newblock


\bibitem[\protect\citeauthoryear{Mikolov, Chen, Corrado, and Dean}{Mikolov
  et~al\mbox{.}}{2013}]%
        {mikolov2013efficient}
\bibfield{author}{\bibinfo{person}{Tomas Mikolov}, \bibinfo{person}{Kai Chen},
  \bibinfo{person}{Greg Corrado}, {and} \bibinfo{person}{Jeffrey Dean}.}
  \bibinfo{year}{2013}\natexlab{}.
\newblock \showarticletitle{Efficient Estimation of Word Representations in
  Vector Space}.
\newblock \bibinfo{journal}{\emph{arXiv preprint arXiv:1301.3781}}
  (\bibinfo{year}{2013}).
\newblock


\bibitem[\protect\citeauthoryear{Myklebust, Jim{\'{e}}nez{-}Ruiz, Chen, Wolf,
  and Tollefsen}{Myklebust et~al\mbox{.}}{2019}]%
        {myklebust2019knowledge}
\bibfield{author}{\bibinfo{person}{Erik~B. Myklebust}, \bibinfo{person}{Ernesto
  Jim{\'{e}}nez{-}Ruiz}, \bibinfo{person}{Jiaoyan Chen}, \bibinfo{person}{Raoul
  Wolf}, {and} \bibinfo{person}{Knut~Erik Tollefsen}.}
  \bibinfo{year}{2019}\natexlab{}.
\newblock \showarticletitle{Knowledge Graph Embedding for Ecotoxicological
  Effect Prediction}.
\newblock \bibinfo{journal}{\emph{CoRR}}  \bibinfo{volume}{abs/1907.01328}.
\newblock
\showeprint[arxiv]{1907.01328}
\urldef\tempurl%
\url{http://arxiv.org/abs/1907.01328}
\showURL{%
\tempurl}


\bibitem[\protect\citeauthoryear{Navarro}{Navarro}{2001}]%
        {navarro2001guided}
\bibfield{author}{\bibinfo{person}{Gonzalo Navarro}.}
  \bibinfo{year}{2001}\natexlab{}.
\newblock \showarticletitle{A Guided Tour to Approximate String Matching}.
\newblock \bibinfo{journal}{\emph{Comput. Surveys}} \bibinfo{volume}{33},
  \bibinfo{number}{1} (\bibinfo{year}{2001}), \bibinfo{pages}{31--88}.
\newblock


\bibitem[\protect\citeauthoryear{Ngomo, Sherif, and Lyko}{Ngomo
  et~al\mbox{.}}{2014}]%
        {ngomo2014unsupervised}
\bibfield{author}{\bibinfo{person}{Axel-Cyrille~Ngonga Ngomo},
  \bibinfo{person}{Mohamed~Ahmed Sherif}, {and} \bibinfo{person}{Klaus Lyko}.}
  \bibinfo{year}{2014}\natexlab{}.
\newblock \showarticletitle{Unsupervised Link Discovery through Knowledge Base
  Repair}. In \bibinfo{booktitle}{\emph{European Semantic Web Conference}}.
  Springer, \bibinfo{pages}{380--394}.
\newblock


\bibitem[\protect\citeauthoryear{Niklaus, Cetto, Freitas, and
  Handschuh}{Niklaus et~al\mbox{.}}{2018}]%
        {niklaus2018survey}
\bibfield{author}{\bibinfo{person}{Christina Niklaus},
  \bibinfo{person}{Matthias Cetto}, \bibinfo{person}{Andr{\'e} Freitas}, {and}
  \bibinfo{person}{Siegfried Handschuh}.} \bibinfo{year}{2018}\natexlab{}.
\newblock \showarticletitle{A Survey on Open Information Extraction}. In
  \bibinfo{booktitle}{\emph{Proceedings of the 27th International Conference on
  Computational Linguistics}}. \bibinfo{pages}{3866--3878}.
\newblock


\bibitem[\protect\citeauthoryear{Niu, Sun, Wang, Rong, Qi, and Yu}{Niu
  et~al\mbox{.}}{2011}]%
        {niu2011zhishi}
\bibfield{author}{\bibinfo{person}{Xing Niu}, \bibinfo{person}{Xinruo Sun},
  \bibinfo{person}{Haofen Wang}, \bibinfo{person}{Shu Rong},
  \bibinfo{person}{Guilin Qi}, {and} \bibinfo{person}{Yong Yu}.}
  \bibinfo{year}{2011}\natexlab{}.
\newblock \showarticletitle{Zhishi.me - Weaving Chinese Linking Open Data}. In
  \bibinfo{booktitle}{\emph{International Semantic Web Conference}}. Springer,
  \bibinfo{pages}{205--220}.
\newblock


\bibitem[\protect\citeauthoryear{Omran, Wang, and Wang}{Omran
  et~al\mbox{.}}{2018}]%
        {omran2018scalable}
\bibfield{author}{\bibinfo{person}{Pouya~Ghiasnezhad Omran},
  \bibinfo{person}{Kewen Wang}, {and} \bibinfo{person}{Zhe Wang}.}
  \bibinfo{year}{2018}\natexlab{}.
\newblock \showarticletitle{Scalable Rule Learning via Learning
  Representation.}. In \bibinfo{booktitle}{\emph{Proceedings of the 27th
  International Joint Conference on Artificial Intelligence}}.
  \bibinfo{pages}{2149--2155}.
\newblock


\bibitem[\protect\citeauthoryear{Paulheim}{Paulheim}{2017}]%
        {paulheim2017knowledge}
\bibfield{author}{\bibinfo{person}{Heiko Paulheim}.}
  \bibinfo{year}{2017}\natexlab{}.
\newblock \showarticletitle{Knowledge Graph Refinement: A Survey of Approaches
  and Evaluation Methods}.
\newblock \bibinfo{journal}{\emph{Semantic web}} \bibinfo{volume}{8},
  \bibinfo{number}{3} (\bibinfo{year}{2017}), \bibinfo{pages}{489--508}.
\newblock


\bibitem[\protect\citeauthoryear{Paulheim and Bizer}{Paulheim and
  Bizer}{2014}]%
        {paulheim2014improving}
\bibfield{author}{\bibinfo{person}{Heiko Paulheim} {and}
  \bibinfo{person}{Christian Bizer}.} \bibinfo{year}{2014}\natexlab{}.
\newblock \showarticletitle{Improving the Quality of Linked Data Using
  Statistical Distributions}.
\newblock \bibinfo{journal}{\emph{International Journal on Semantic Web and
  Information Systems (IJSWIS)}} \bibinfo{volume}{10}, \bibinfo{number}{2}
  (\bibinfo{year}{2014}), \bibinfo{pages}{63--86}.
\newblock


\bibitem[\protect\citeauthoryear{Paulheim and Gangemi}{Paulheim and
  Gangemi}{2015}]%
        {paulheim2015serving}
\bibfield{author}{\bibinfo{person}{Heiko Paulheim} {and} \bibinfo{person}{Aldo
  Gangemi}.} \bibinfo{year}{2015}\natexlab{}.
\newblock \showarticletitle{Serving DBpedia with DOLCE -- More than Just Adding
  A Cherry on Top}. In \bibinfo{booktitle}{\emph{International Semantic Web
  Conference}}. Springer, \bibinfo{pages}{180--196}.
\newblock


\bibitem[\protect\citeauthoryear{Pellissier~Tanon, Bourgaux, and
  Suchanek}{Pellissier~Tanon et~al\mbox{.}}{2019}]%
        {pellissier2019learning}
\bibfield{author}{\bibinfo{person}{Thomas Pellissier~Tanon},
  \bibinfo{person}{Camille Bourgaux}, {and} \bibinfo{person}{Fabian Suchanek}.}
  \bibinfo{year}{2019}\natexlab{}.
\newblock \showarticletitle{Learning How to Correct A Knowledge Base from the
  Edit History}. In \bibinfo{booktitle}{\emph{The World Wide Web Conference}}.
  ACM, \bibinfo{pages}{1465--1475}.
\newblock


\bibitem[\protect\citeauthoryear{Pujara, Augustine, and Getoor}{Pujara
  et~al\mbox{.}}{2017}]%
        {pujara2017sparsity}
\bibfield{author}{\bibinfo{person}{Jay Pujara}, \bibinfo{person}{Eriq
  Augustine}, {and} \bibinfo{person}{Lise Getoor}.}
  \bibinfo{year}{2017}\natexlab{}.
\newblock \showarticletitle{Sparsity and Noise: Where Knowledge Graph
  Embeddings Fall Short}. In \bibinfo{booktitle}{\emph{Proceedings of the 2017
  Conference on Empirical Methods in Natural Language Processing}}.
  \bibinfo{pages}{1751--1756}.
\newblock


\bibitem[\protect\citeauthoryear{Ristoski and Paulheim}{Ristoski and
  Paulheim}{2016}]%
        {ristoski2016rdf2vec}
\bibfield{author}{\bibinfo{person}{Petar Ristoski} {and} \bibinfo{person}{Heiko
  Paulheim}.} \bibinfo{year}{2016}\natexlab{}.
\newblock \showarticletitle{RDF2Vec: RDF Graph Embeddings for Data Mining}. In
  \bibinfo{booktitle}{\emph{International Semantic Web Conference}}. Springer,
  \bibinfo{pages}{498--514}.
\newblock


\bibitem[\protect\citeauthoryear{Socher, Chen, Manning, and Ng}{Socher
  et~al\mbox{.}}{2013}]%
        {socher2013reasoning}
\bibfield{author}{\bibinfo{person}{Richard Socher}, \bibinfo{person}{Danqi
  Chen}, \bibinfo{person}{Christopher~D Manning}, {and} \bibinfo{person}{Andrew
  Ng}.} \bibinfo{year}{2013}\natexlab{}.
\newblock \showarticletitle{Reasoning with Neural Tensor Networks for Knowledge
  Base Completion}. In \bibinfo{booktitle}{\emph{Advances in Neural Information
  Processing Systems}}. \bibinfo{pages}{926--934}.
\newblock


\bibitem[\protect\citeauthoryear{Tonon, Catasta, Demartini, and
  Cudr{\'e}-Mauroux}{Tonon et~al\mbox{.}}{2015}]%
        {tonon2015fixing}
\bibfield{author}{\bibinfo{person}{Alberto Tonon}, \bibinfo{person}{Michele
  Catasta}, \bibinfo{person}{Gianluca Demartini}, {and}
  \bibinfo{person}{Philippe Cudr{\'e}-Mauroux}.}
  \bibinfo{year}{2015}\natexlab{}.
\newblock \showarticletitle{Fixing the Domain and Range of Properties in Linked
  Data by Context Disambiguation.}
\newblock \bibinfo{journal}{\emph{LDOW@ WWW}}  \bibinfo{volume}{1409}
  (\bibinfo{year}{2015}).
\newblock


\bibitem[\protect\citeauthoryear{T{\"o}pper, Knuth, and Sack}{T{\"o}pper
  et~al\mbox{.}}{2012}]%
        {topper2012dbpedia}
\bibfield{author}{\bibinfo{person}{Gerald T{\"o}pper}, \bibinfo{person}{Magnus
  Knuth}, {and} \bibinfo{person}{Harald Sack}.}
  \bibinfo{year}{2012}\natexlab{}.
\newblock \showarticletitle{DBpedia Ontology Enrichment for Inconsistency
  Detection}. In \bibinfo{booktitle}{\emph{Proceedings of the 8th International
  Conference on Semantic Systems}}. ACM, \bibinfo{pages}{33--40}.
\newblock


\bibitem[\protect\citeauthoryear{Toutanova and Chen}{Toutanova and
  Chen}{2015}]%
        {toutanova2015observed}
\bibfield{author}{\bibinfo{person}{Kristina Toutanova} {and}
  \bibinfo{person}{Danqi Chen}.} \bibinfo{year}{2015}\natexlab{}.
\newblock \showarticletitle{Observed versus Latent Features for Knowledge Base
  and Text Inference}. In \bibinfo{booktitle}{\emph{Proceedings of the 3rd
  Workshop on Continuous Vector Space Models and their Compositionality}}.
  \bibinfo{pages}{57--66}.
\newblock


\bibitem[\protect\citeauthoryear{Vashishth, Jain, and Talukdar}{Vashishth
  et~al\mbox{.}}{2018}]%
        {vashishth2018cesi}
\bibfield{author}{\bibinfo{person}{Shikhar Vashishth}, \bibinfo{person}{Prince
  Jain}, {and} \bibinfo{person}{Partha Talukdar}.}
  \bibinfo{year}{2018}\natexlab{}.
\newblock \showarticletitle{CESI: Canonicalizing Open Knowledge Bases Using
  Embeddings and Side Information}. In \bibinfo{booktitle}{\emph{Proceedings of
  the 2018 World Wide Web Conference}}. International World Wide Web
  Conferences Steering Committee, \bibinfo{pages}{1317--1327}.
\newblock


\bibitem[\protect\citeauthoryear{Vrande{\v{c}}i{\'c} and
  Kr{\"o}tzsch}{Vrande{\v{c}}i{\'c} and Kr{\"o}tzsch}{2014}]%
        {vrandevcic2014wikidata}
\bibfield{author}{\bibinfo{person}{Denny Vrande{\v{c}}i{\'c}} {and}
  \bibinfo{person}{Markus Kr{\"o}tzsch}.} \bibinfo{year}{2014}\natexlab{}.
\newblock \showarticletitle{Wikidata: A Free Collaborative Knowledge Base}.
\newblock  (\bibinfo{year}{2014}).
\newblock


\bibitem[\protect\citeauthoryear{Wang, Mao, Wang, and Guo}{Wang
  et~al\mbox{.}}{2017}]%
        {wang2017knowledge}
\bibfield{author}{\bibinfo{person}{Quan Wang}, \bibinfo{person}{Zhendong Mao},
  \bibinfo{person}{Bin Wang}, {and} \bibinfo{person}{Li Guo}.}
  \bibinfo{year}{2017}\natexlab{}.
\newblock \showarticletitle{Knowledge Graph Embedding: A Survey of Approaches
  and Applications}.
\newblock \bibinfo{journal}{\emph{IEEE Transactions on Knowledge and Data
  Engineering}} \bibinfo{volume}{29}, \bibinfo{number}{12}
  (\bibinfo{year}{2017}), \bibinfo{pages}{2724--2743}.
\newblock


\bibitem[\protect\citeauthoryear{Weaver, Strickland, and Crane}{Weaver
  et~al\mbox{.}}{2006}]%
        {weaver2006quantifying}
\bibfield{author}{\bibinfo{person}{Gabriel Weaver}, \bibinfo{person}{Barbara
  Strickland}, {and} \bibinfo{person}{Gregory Crane}.}
  \bibinfo{year}{2006}\natexlab{}.
\newblock \showarticletitle{Quantifying the Accuracy of Relational Statements
  in Wikipedia: A Methodology}. In \bibinfo{booktitle}{\emph{Proceedings of the
  6th ACM/IEEE-CS Joint Conference on Digital Libraries}},
  Vol.~\bibinfo{volume}{6}. Citeseer, \bibinfo{pages}{358--358}.
\newblock


\bibitem[\protect\citeauthoryear{Wu, Wu, Kao, and Yin}{Wu
  et~al\mbox{.}}{2018}]%
        {wu2018towards}
\bibfield{author}{\bibinfo{person}{Tien-Hsuan Wu}, \bibinfo{person}{Zhiyong
  Wu}, \bibinfo{person}{Ben Kao}, {and} \bibinfo{person}{Pengcheng Yin}.}
  \bibinfo{year}{2018}\natexlab{}.
\newblock \showarticletitle{Towards Practical Open Knowledge Base
  Canonicalization}. In \bibinfo{booktitle}{\emph{Proceedings of the 27th ACM
  International Conference on Information and Knowledge Management}}. ACM,
  \bibinfo{pages}{883--892}.
\newblock


\bibitem[\protect\citeauthoryear{Yang, Yih, He, Gao, and Deng}{Yang
  et~al\mbox{.}}{2014}]%
        {yang2015embedding}
\bibfield{author}{\bibinfo{person}{Bishan Yang}, \bibinfo{person}{Wen-tau Yih},
  \bibinfo{person}{Xiaodong He}, \bibinfo{person}{Jianfeng Gao}, {and}
  \bibinfo{person}{Li Deng}.} \bibinfo{year}{2014}\natexlab{}.
\newblock \showarticletitle{Embedding Entities and Relations for Learning and
  Inference in Knowledge Bases}.
\newblock \bibinfo{journal}{\emph{arXiv preprint arXiv:1412.6575}}
  (\bibinfo{year}{2014}).
\newblock


\bibitem[\protect\citeauthoryear{Zhang, Paudel, Wang, Chen, Zhu, Zhang,
  Bernstein, and Chen}{Zhang et~al\mbox{.}}{2019}]%
        {zhang2019iteratively}
\bibfield{author}{\bibinfo{person}{Wen Zhang}, \bibinfo{person}{Bibek Paudel},
  \bibinfo{person}{Liang Wang}, \bibinfo{person}{Jiaoyan Chen},
  \bibinfo{person}{Hai Zhu}, \bibinfo{person}{Wei Zhang},
  \bibinfo{person}{Abraham Bernstein}, {and} \bibinfo{person}{Huajun Chen}.}
  \bibinfo{year}{2019}\natexlab{}.
\newblock \showarticletitle{Iteratively Learning Embeddings and Rules for
  Knowledge Graph Reasoning}. In \bibinfo{booktitle}{\emph{The World Wide Web
  Conference}}. ACM, \bibinfo{pages}{2366--2377}.
\newblock


\end{thebibliography}

\end{document}